\pdfoutput=1
\documentclass[letterpaper]{article}

\usepackage[preprint]{aaai2027}

\usepackage[hyphens]{url}
\usepackage{graphicx}
\usepackage{natbib}
\usepackage{caption}
\usepackage{booktabs}
\usepackage{array}
\usepackage[table]{xcolor}
\usepackage{amsmath,amssymb,mathtools}
\usepackage{multirow}
\usepackage{makecell}
\usepackage{threeparttable}
\usepackage{siunitx}
\usepackage[accsupp]{axessibility}
\usepackage{pifont}
\usepackage{placeins}
\usepackage{subcaption}

\newcommand{\PM}{\ensuremath{\text{\textsc{PM}}}}
\newcommand{\PU}{\ensuremath{\text{\textsc{PU}}}}
\newcommand{\CEOC}{\ensuremath{\text{\textsc{CursiveEOC}}}}
\newcommand{\EOC}{\ensuremath{\text{\textsc{EOC}}}}

\definecolor{YesCell}{RGB}{212,235,210}
\definecolor{NoCell}{RGB}{235,206,206}
\definecolor{PartCell}{RGB}{236,232,191}

\newcolumntype{L}[1]{>{\raggedright\arraybackslash}m{#1}}
\newcolumntype{C}[1]{>{\centering\arraybackslash}m{#1}}

\newcommand{\resulttablefont}{%
  \fontsize{8.0pt}{9.2pt}\selectfont
}

\newcommand{\suppname}{appendix}

\newcommand{\suppref}[1]{Appendix~\ref{#1}}

\newcommand{\suppsee}[1]{see Appendix~\ref{#1}}
\begin{document}

% ===============================================================
% Title
% ===============================================================

\title{BoundInk: Boundary-Aware Online Handwriting Generation}

% ===============================================================
% Authors
% ===============================================================

\author{
    Jinsu Shin,
    Sungeun Hong\textsuperscript{*},
    JinYeong Bak\textsuperscript{*}
}

\affiliations{
    Sungkyunkwan University, Suwon, Korea\\
    \textsuperscript{*}Corresponding authors: csehong@skku.edu, jy.bak@skku.edu
}

\maketitle

\begin{abstract}
Realistic online handwriting depends not only on individual character shapes, but also on how a writer connects, spaces, and aligns adjacent characters. Existing methods rely primarily on long-range sequence modeling and capture these inter-character behaviors only implicitly. This often produces plausible glyphs accompanied by broken cursive joins, inconsistent spacing, or writer-inconsistent transitions. We introduce \textsc{BoundInk}, a writer-conditioned framework that treats inter-character boundaries as explicit generation units. By jointly modeling local transitions and surrounding text context, \textsc{BoundInk} preserves writer-specific glyph appearance while improving connectivity and spacing across complete text lines. We further introduce a boundary-aware evaluation framework that directly assesses cursive continuity and spatial relationships between characters beyond conventional trajectory similarity. Across three benchmark-matched settings, \textsc{BoundInk} improves all applicable boundary-quality measures and reduces normalized dynamic time warping by 17.6--47.8\%. In blind human evaluations, BoundInk outputs are preferred in 78.0--82.6\% of valid criterion-wise judgments.

\end{abstract}

%\footnote{Code and Data: \url{https://github.com/jinsu0000/BoundInk-official}}

%\keywords{Online Handwriting \and Sentence Generation \and Boundary-Aware Generation \and Inter-Character Connectivity \and Spacing}
\section{Introduction}
Online handwriting represents writing as a time-ordered sequence of coordinates and pen states~\cite{faundez2021online}.
Offline handwriting, in contrast, records only the final static image~\cite{plamondon2000online}.
Sentence-level generation must therefore recover both the appearance and temporal dynamics of handwriting~\cite{ott2022benchmarking}.
These dynamics include stroke order, pen-state transitions, and interactions between neighboring characters.
Natural handwriting further requires writer-consistent connections, spacing, and alignment across character boundaries.

\begin{figure}[t!]
    \centering
    \includegraphics[width=\columnwidth]{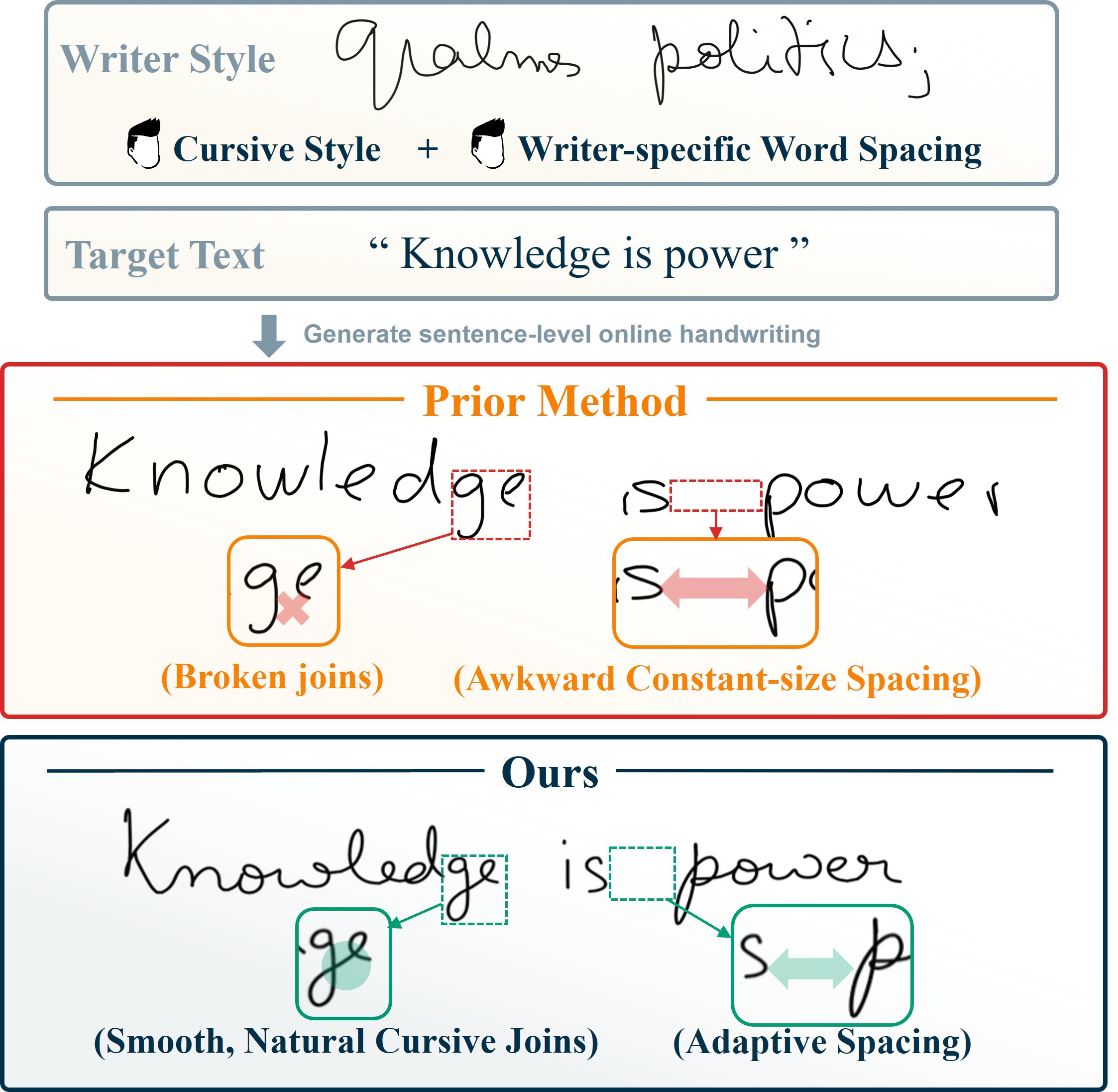}
\caption{
Sentence-level handwriting requires both faithful glyphs and natural boundary behavior.
Prior methods often produce broken cursive connections and rigid, writer-agnostic spacing.
BoundInk instead generates smooth character transitions and adaptive writer-specific spacing.
}
    \label{fig:intro_teaser}
\end{figure}

Most existing approaches generate an entire sentence through long-range sequential modeling~\cite{graves2013generating,aksan2018deepwriting,kotani2020generating,jin2025trink}.
Writer-conditioned methods additionally extract style from reference trajectories or writer descriptors~\cite{aksan2018deepwriting,kotani2020generating}.
Although these approaches capture overall appearance, inter-character transitions remain implicit.
Consequently, they can produce recognizable glyphs with broken cursive joins, rigid spacing, or writer-inconsistent transitions~\cite{ren2025decoupling,kotani2020generating,aksan2018deepwriting}.
As illustrated in Figure~\ref{fig:intro_teaser}, writer style is expressed both within glyphs and across their connections and spatial relationships.
We refer to these inter-character properties as \emph{boundary behavior}, with key terms summarized in Table~\ref{tab:terminology}.

\begin{table}[t]
\centering
\small
\setlength{\tabcolsep}{4pt}
\begin{tabular}{p{0.3\columnwidth} p{0.62\columnwidth}}
\toprule
\textbf{Term} & \textbf{Definition} \\
\midrule
Glyph &
The writer-conditioned trajectory realization of a character \\
Inter-character boundary &
The transition between two adjacent characters. \\
Boundary behavior &
Connectivity, spacing, and relative alignment at a boundary \\
Cursive connection &
Pen-down continuation across a character boundary \\
Inter-character spacing &
The horizontal separation or overlap between adjacent non-space glyphs \\
Word spacing &
The horizontal separation induced by a whitespace boundary \\
\bottomrule
\end{tabular}
\caption{Key terminology used in this work.}
\label{tab:terminology}
\end{table}

In this study, we introduce \textsc{BoundInk}, a boundary-aware framework for writer-conditioned online handwriting generation.
The key idea is to treat inter-character boundaries as explicit generation targets rather than incidental outcomes of long-sequence decoding.
\textsc{BoundInk} models each predecessor--current transition locally and supplements it with position-specific sentence context.
This division captures immediate boundary dynamics without discarding information from the surrounding sentence.
It enables the model to preserve writer-specific glyph appearance while adapting cursive connections, spacing, and alignment to both neighboring characters and sentence context.

Boundary quality is also difficult to assess using conventional trajectory distances alone.
A globally similar trajectory can still contain incorrect connections or unnatural spacing.
We therefore introduce \emph{Connectivity and Spacing Metrics (CSM)}, a writer-level evaluation suite that directly measures cursive connectivity, inter-character spacing, and word spacing.
Across three benchmark-matched protocols, \textsc{BoundInk} improves all applicable boundary-aware measures and reduces normalized trajectory distance by 17.6--47.8\%.
It also achieves stronger writer-style preservation while maintaining competitive content legibility.
Blind human evaluations further prefer \textsc{BoundInk} in 78.0--82.6\% of valid criterion-wise judgments, confirming improvements in connectivity, spacing, and writer-style fidelity.

\smallskip
\noindent Our contributions are summarized as follows:
\begin{itemize}
    \item \textbf{Explicit modeling of inter-character boundaries.}
    We introduce \textsc{BoundInk}, which elevates inter-character transitions from implicit sequence artifacts to explicit generation targets.
    It integrates local predecessor--current dynamics with sentence context to generate writer-consistent glyphs, connections, spacing, and alignment.

    \item \textbf{Boundary-aware evaluation.}
    We introduce \emph{Connectivity and Spacing Metrics}, which directly assess cursive connectivity and spatial relationships between adjacent characters.
    These measures complement global trajectory similarity with targeted evaluation of sentence-level boundary quality.

    \item \textbf{Comprehensive empirical validation.}
    Across three benchmark-matched settings, \textsc{BoundInk} consistently improves all applicable boundary-aware measures and normalized trajectory similarity.
    Human preference results further demonstrate improvements in perceived connectivity, spacing, and writer-style fidelity.
\end{itemize}
\section{Related Work}
\label{sec:related}

\begin{table}[t!]
\centering
\small
\setlength{\tabcolsep}{1.5pt}
\renewcommand{\arraystretch}{0.95}

\begin{tabular*}{\columnwidth}{@{\extracolsep{\fill}}lcccc@{}}
\toprule
\textbf{Property} &
DeepWriting &
DSD &
OLHWG &
\textbf{Ours} \\
\midrule
Word-spacing modeling
    & $\triangle$ & \textemdash & $\checkmark$ & $\checkmark$ \\
Boundary supervision
    & \textemdash & \textemdash & \textemdash & $\checkmark$ \\
Image-based style ref.
    & \textemdash & \textemdash & \textemdash & $\checkmark$ \\
\bottomrule
\end{tabular*}
\caption{Design properties of benchmark-matched methods used in our sentence-level comparisons.
$\checkmark$: explicit support; $\triangle$: partial or implicit support; \textemdash: no dedicated mechanism.}
\label{tab:capability_comparison}
\end{table}

\subsection{Personalized Writer Stylization}
\label{sec:rw_char_identity}

Personalized online handwriting generation aims to preserve target content while reproducing a writer's style from limited references.
Prior methods improve controllability by separating writer-dependent style from character or glyph content~\cite{tang2021write,kotani2020generating,dai2023disentangling,zhao2020deep}.
\emph{Elegantly Written}~\cite{liu2024elegantly} disentangles writer and character styles for online Chinese handwriting enhancement, but operates on observed character trajectories rather than sentence-level generation.
Offline handwriting generators similarly separate style and content~\cite{kang2020ganwriting,gan2021higan,luo2022slogan}, yet produce static images instead of dynamic pen trajectories.
These approaches primarily represent style through individual glyph appearance.
\textsc{BoundInk} additionally captures how writer style is expressed between neighboring characters, including their connections, spacing, and alignment.

\subsection{Inter-Character Boundary Modeling}
\label{sec:rw_connectivity}

Sentence-level realism depends on how adjacent characters connect and align. These boundary properties include cursive continuation, inter-character spacing, and relative alignment~\cite{nakatsuru2024learning,plamondon2000online,lee2010over}.
Most neural generators learn these boundary properties only implicitly through sequence modeling~\cite{graves2013generating,zhang2017drawing,tang2019fontrnn,tolosana2021deepwritesyn,bhunia2021handwriting}, often producing broken joins or unstable spacing.
DeepWriting~\cite{aksan2018deepwriting} and DSD~\cite{kotani2020generating} improve personalized trajectory generation without explicit boundary supervision.
OLHWG~\cite{ren2025decoupling} models layout and spacing, but not pen-down continuation across characters.
In contrast, we treat inter-character transitions as explicit generation targets and jointly model local boundary dynamics with sentence context. Table~\ref{tab:capability_comparison} summarizes these differences
under the benchmark-matched comparison settings.
\begin{figure*}[!t]
    \centering
    \includegraphics[
        width=0.94\textwidth,
        % trim=0 2mm 0 0.5mm,
        % clip
    ]{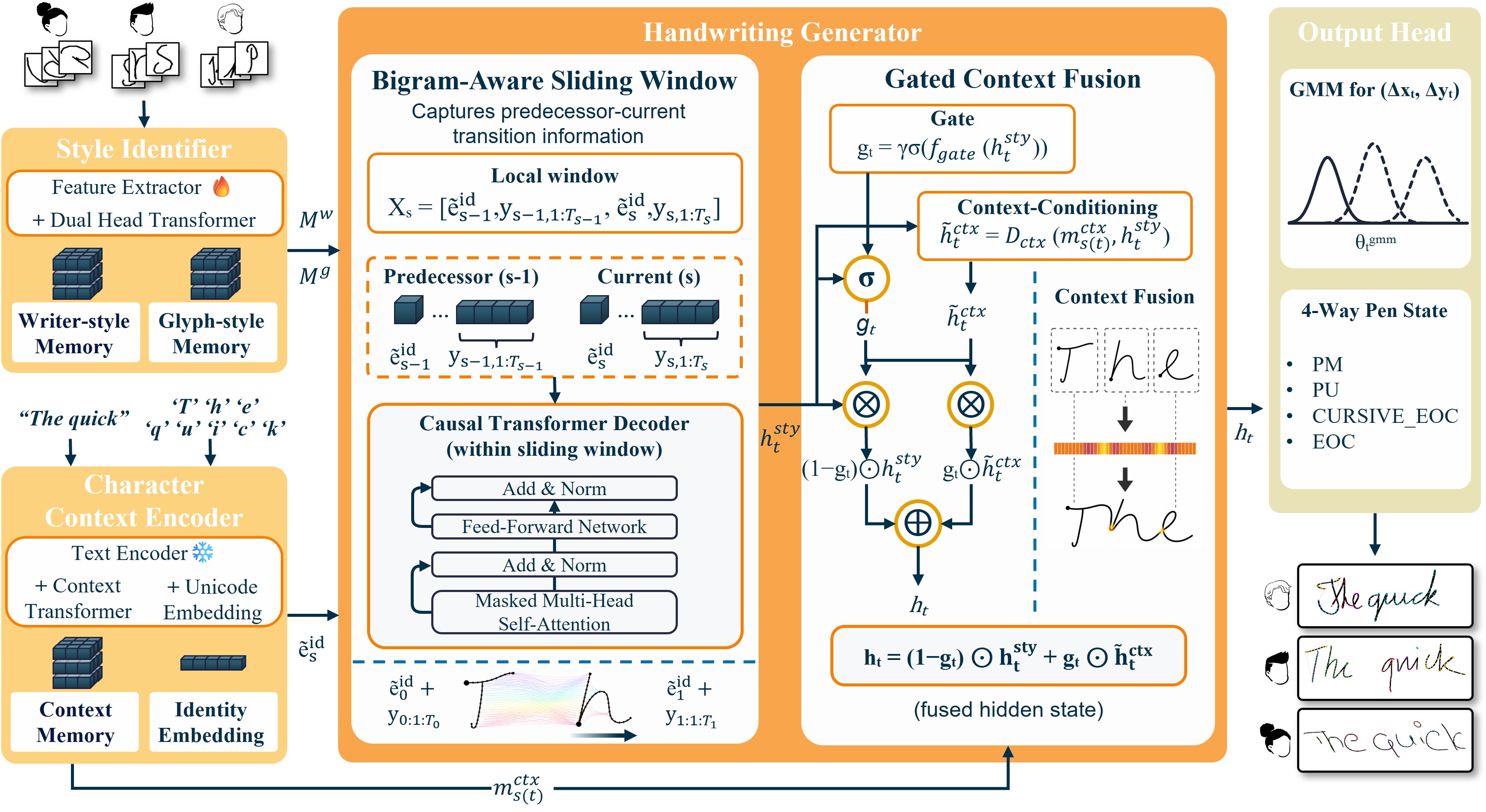}
    \caption{
    \textsc{BoundInk} overview. % of \textsc{BoundInk}.
    The framework (1) extracts writer- and glyph-style cues from reference handwriting,
    (2) encodes character identity and sentence context from target text, and
    (3) generates each character from a predecessor--current window with gated context fusion.
    Resulting trajectories preserve writer-specific appearance while adapting connectivity and spacing.
    }
    \label{fig:method_overview}
\end{figure*}

\subsection{Limited Transition Diversity}
\label{sec:rw_composition}

Sentence-level datasets cover only a subset of possible character transitions.
Rare character pairs therefore remain difficult to generate reliably, even when individual glyphs are accurate.
Recurrent models capture long-range trajectory dependencies~\cite{graves2013generating,aksan2018deepwriting}, while Transformer-based approaches strengthen full-sequence modeling~\cite{bhunia2021handwriting,jin2025trink}.
Layout-oriented methods instead separate spatial arrangement from glyph generation~\cite{ren2025decoupling}.
However, local boundary learning remains coupled with complete sentence generation.
\textsc{BoundInk} separates these roles by modeling predecessor--current transitions locally and supplying longer-range information through sentence context.
This design emphasizes reusable local boundary patterns while preserving sentence-level coherence.
\section{Method}
%\section{BoundInk}
\label{sec:method}

We aim to generate writer-conditioned online handwriting while explicitly modeling transitions between adjacent characters.
The central design separates four complementary sources of information:
(i) \emph{what to write}, represented by character identity;
(ii) \emph{whose handwriting to imitate}, represented by global writer style;
(iii) \emph{how each character should be realized}, represented by glyph-specific style; and
(iv) \emph{how neighboring characters should interact}, represented by sentence context and local boundary history.

As shown in Figure~\ref{fig:method_overview}, the Style Identifier extracts writer and glyph styles from reference handwriting, while the Character Context Encoder represents the target text. A bigram-aware sliding-window Transformer then generates each character from its predecessor and surrounding sentence context. This decomposition captures the core intuition of \textsc{BoundInk}: character appearance should remain writer-consistent, while connections, spacing, and alignment adapt to neighboring characters and sentence position. Subscripts denote sequence indices such as $s$ and $t$, superscripts denote representation types such as $\mathrm{id}$, $\mathrm{ctx}$, and $\mathrm{sty}$, and roman subscripts denote modules such as $f_{\mathrm{text}}$ and $f_{\mathrm{ctx}}$.

\subsection{Character Identity and Context Fusion}
\label{subsec:context_encoder}
The Character Context Encoder separates target-text information into two signals for sentence-level generation.
A \emph{Character-Identity Embedding} specifies \emph{which character to draw}, while position-dependent \emph{context memory} modulates \emph{how it should appear and connect} within the sentence.
Both signals are derived from a shared multilingual text encoder $f_{\mathrm{text}}$ (e.g., CANINE~\cite{clark2022canine} or ByT5~\cite{xue2022byt5}), and the context branch adds a lightweight Transformer encoder for sentence-dependent variation.
For a Unicode character $u$, we encode it in isolation to avoid contextual leakage and obtain a deterministic identity anchor \(\mathbf{e}^{\mathrm{id}}(u)=f_{\mathrm{id}}(f_{\mathrm{text}}([u]))\in\mathbb{R}^{D}\), where $f_{\mathrm{id}}(\cdot)$ is a pooling-and-projection head.

In practice, $\mathbf{e}^{\mathrm{id}}(u)$ is cached for unique codepoints within a mini-batch for efficient reuse.
To sharpen discrete identity signals, we augment the Character-Identity Embedding as \(\tilde{\mathbf{e}}^{\mathrm{id}}(u)=\mathbf{e}^{\mathrm{id}}(u)+\alpha\,\mathbf{P}\phi(u)\), with \(\alpha\in\mathbb{R}_{+}\), where \(\phi(u)\) is a learnable code embedding, \(\mathbf{P}\) is a projection matrix, and \(\alpha\) is a learnable scalar initialized to a small positive value.
This improves identity separability (e.g., case-sensitive distinctions) while keeping identity specification separate from sentence-dependent modulation.
For a sentence-level character sequence $\mathbf{u}=(u_1,\ldots,u_S)$, we compute sentence-aware context memory as \(\mathbf{H}^{\mathrm{text}}=\mathbf{W}_{h}f_{\mathrm{text}}(\mathbf{u})\) and \(\mathbf{M}^{\mathrm{ctx}}=f_{\mathrm{ctx}}(\mathrm{TransEnc}_{\mathrm{ctx}}(\mathbf{H}^{\mathrm{text}}))\), with $f_{\mathrm{ctx}}(\cdot)$ a projection head.
The resulting vector $\mathbf{m}^{\mathrm{ctx}}_s$ varies with neighboring characters and sentence position, making it suitable for modeling inter-character connectivity (e.g., spacing/kerning and cursive joins).
Context memory helps adapt inter-character connectivity, but excessive context injection can weaken writer/glyph stylization.

Let $\mathbf{c}_t=\mathbf{m}^{\mathrm{ctx}}_{s(t)}$ denote the position-specific context vector for token $t$, and let $\widetilde{\mathbf{h}}^{\mathrm{ctx}}_t$ denote the context-decoder output obtained by conditioning $\mathbf{h}^{\mathrm{sty}}_t$ on $\mathbf{c}_t$.
We therefore use token-wise \emph{gated context fusion}:
\begin{equation}
\small
g_t=\gamma\,\sigma\!\left(f_{\mathrm{gate}}\!\left(\mathbf{h}^{\mathrm{sty}}_t\right)\right),
\quad
\mathbf{h}_t=(1-g_t)\mathbf{h}^{\mathrm{sty}}_t
+g_t\widetilde{\mathbf{h}}^{\mathrm{ctx}}_t ,
\label{eq:gated_context_fusion}
\end{equation}
where $g_t\in(0,\gamma)$ is a scalar gate broadcast over the feature dimension, $\gamma\in(0,1]$ is the gate cap, and $s(t)$ maps token $t$ to its character position.
The gate controls how much sentence context is injected at each token, allowing local connectivity adaptation without overwriting writer/glyph style cues.

\subsection{Bigram-Aware Local Decoding}
\label{subsec:bigram_decoder}

\begin{figure}[t!]
    \centering
    \includegraphics[width=1.0\linewidth]{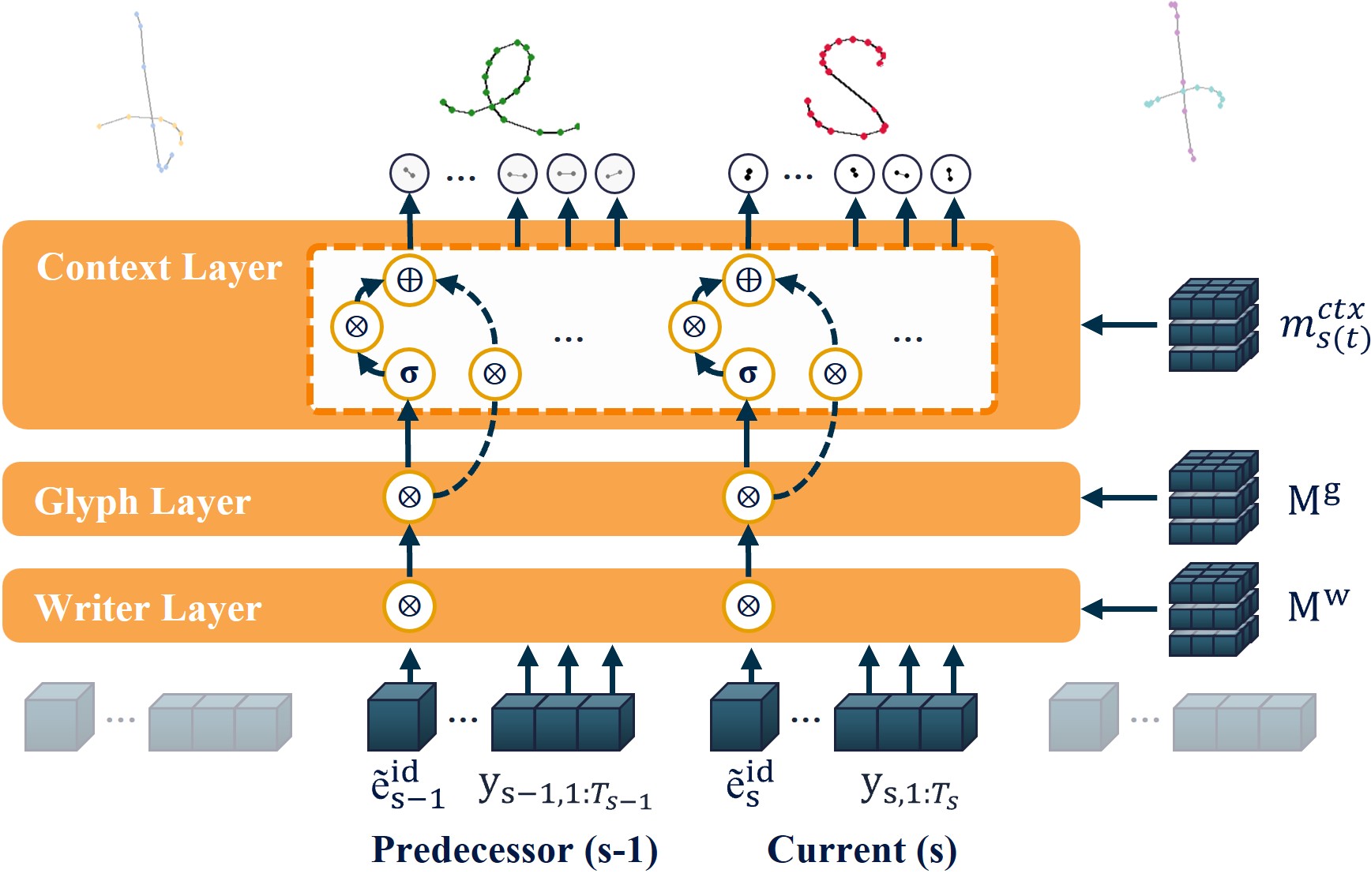}
 \caption{
    Token-wise decoding with the bigram-aware sliding-window Transformer.
    Bi-SWT generates each character from a predecessor--current window using writer- and glyph-style cues.
    Gated context fusion injects sentence-level information to adapt connectivity, spacing, and cursive joins.
    }
    \label{fig:method_decoder_with_bigram}
\end{figure}

\textsc{BoundInk} uses a \emph{bigram-aware sliding-window Transformer decoder} (Bi-SWT decoder) to model predecessor-conditioned local transitions.
Because kerning and cursive joins are primarily determined by adjacent characters, the decoder operates on predecessor--current bigram windows.
This design captures the dominant boundary signal while reducing the compositional sparsity of larger \emph{n}-gram windows.
Longer-range sentence information is supplied separately through \emph{context memory} and \emph{gated context fusion}.
Figure~\ref{fig:method_decoder_with_bigram} illustrates the decoding process.
Character-identity embeddings and trajectory tokens form the local window, while \emph{Writer-style memory} and \emph{Glyph-style memory} provide stylization cues.
Gated context fusion injects position-specific sentence context, allowing boundary transitions to adapt without overriding writer-specific appearance.

For character position $s$, let $\tilde{\mathbf{e}}^{\mathrm{id}}_s$ be the augmented Character-Identity Embedding produced by the Character Context Encoder, and let $\mathbf{y}_{s,1:T_s}$ denote the trajectory-token sequence for the character at position $s$.
During training with teacher forcing, we construct a local window from the predecessor and current characters:
\begin{equation}
\mathbf{X}_s =
[\ \tilde{\mathbf{e}}^{\mathrm{id}}_{s-1},\ \mathbf{y}_{s-1,1:T_{s-1}},\ \tilde{\mathbf{e}}^{\mathrm{id}}_s,\ \mathbf{y}_{s,1:T_s}\ ]
\in \mathbb{R}^{L_s \times D},
\label{eq:bigram_window}
\end{equation}
where $L_s = 2 + T_{s-1} + T_s$.
For $s=1$, the window reduces to the current-character stream only.
At inference time, $\mathbf{y}_{s-1,\cdot}$ is replaced by the model prediction, while the Bi-SWT decoder autoregressively generates the current character trajectory.
Within each local window, the Bi-SWT decoder applies causal self-attention with key-padding masks for variable-length trajectory tokens.
To encode relative ordering inside the predecessor-current window, we apply RoPE~\cite{su2024roformer} to query and key projections.

Each decoder block conditions on two complementary sources: (i) \emph{Writer-style memory} and \emph{Glyph-style memory}, which are encoded from reference handwriting, and (ii) position-specific \emph{context memory}, which is produced by the Character Context Encoder.
Context memory is integrated through \emph{gated context fusion} in Eq.~\eqref{eq:gated_context_fusion}, allowing local adaptation without collapsing writer/glyph stylization.

The decoder predicts point-to-point trajectory deltas autoregressively rather than absolute coordinates.
For each valid step $t$, the decoder hidden state $\mathbf{h}_t$ is mapped to an output head predicting bivariate $K$-component Gaussian Mixture Model (GMM) parameters over \((\Delta x_t,\Delta y_t)\) and 4-way pen-state logits. 
We use $K=20$, and the head outputs \((\boldsymbol{\theta}^{\mathrm{gmm}}_{t},\boldsymbol{\ell}^{\mathrm{pen}}_{t})=f_{\mathrm{out}}(\mathbf{h}_t)\).
We obtain mixture parameters using standard squashing functions (e.g., softmax for mixture weights, exponential for standard deviations, and tanh for correlations).
Pen-state predictions are supervised by connectivity-aware boundary labels in the training objective.

\subsection{Style and Text Conditioning}
\label{subsec:style_content}

\textsc{BoundInk} separates style conditioning from text-side conditioning. Following the SDT style-reference protocol~\cite{dai2023disentangling}, the style pathway encodes reference handwriting images into \emph{Writer-style memory} and \emph{Glyph-style memory}, providing global writer traits and character-specific style cues for writer-consistent decoding. The text-side pathway uses a \emph{Character Context Encoder}, where Unicode-based \emph{Character-Identity Embeddings} replace font-rendered target-content images, reducing inference-time input overhead.

\subsection{Curriculum and Boundary Supervision}
\label{subsec:training_objectives}

We train \textsc{BoundInk} with a three-stage curriculum, from glyphs to bigrams and then full sentences.
The curriculum first learns writer-consistent glyph formation, then predecessor-current transitions, and finally sentence-level composition.
This exposes the model to connectivity-related boundary events before long sentence generation.
At stage $k$, the active generation loss is
\begin{equation}
\mathcal{L}_{\mathrm{gen}}^{(k)}
=
\lambda_{\mathrm{char}}^{(k)}\mathcal{L}_{\mathrm{char}}
+
\lambda_{\mathrm{bi}}^{(k)}\mathcal{L}_{\mathrm{bi}}
+
\lambda_{\mathrm{sent}}^{(k)}\mathcal{L}_{\mathrm{sent}},
\label{eq:curriculum_loss}
\end{equation}
where the three terms correspond to representative glyph samples, predecessor-current bigram windows, and sentence sliding windows, respectively.
Using the decoder outputs, we optimize a masked GMM negative log-likelihood on valid trajectory steps:
\begin{equation}
\mathcal{L}_{\mathrm{mdn}}
=
-\frac{1}{|\mathcal{V}|}
\sum_{t\in\mathcal{V}}
\log\!\left(
\sum_{k=1}^{K}
\pi_{t,k}\,
\mathcal{N}_2\!\left(\Delta\mathbf{y}_t;\boldsymbol{\mu}_{t,k},\boldsymbol{\Sigma}_{t,k}\right)
\right),
\label{eq:gmm_nll}
\end{equation}
where $\mathcal{V}$ is the set of valid (non-padded) steps, $\Delta\mathbf{y}_t=(\Delta x_t,\Delta y_t)$, and $\pi_{t,k}$ are mixture weights.

We also apply masked cross-entropy on 4-way pen states (\textsc{PM}, \textsc{PU}, \textsc{CursiveEOC}, \textsc{EOC}) as \(\mathcal{L}_{\mathrm{pen}}=-\frac{1}{|\mathcal{V}|}\sum_{t\in\mathcal{V}}\log p^{\mathrm{pen}}_{t,\,y^{\mathrm{pen}}_t}\), where \textsc{CursiveEOC} explicitly supervises pen-down continuation across character boundaries (cursive joins).
All pen state labels, including \textsc{CursiveEOC}, are deterministically derived from the preprocessed trajectory stream and boundary annotations.
We define the sequence loss as \(\mathcal{L}_{\mathrm{seq}}=\mathcal{L}_{\mathrm{mdn}}+\lambda_{\mathrm{pen}}\mathcal{L}_{\mathrm{pen}}\), and use it for both glyph and sentence streams.
To preserve a unified trajectory formulation, we retain inter-word spaces as short pseudo-trajectory segments rather than dropping them.

For the bigram stream, we add a lightweight \emph{Vertical Drift Loss (VDL)} as an auxiliary regularizer to reduce predecessor-current boundary misalignment, especially \emph{y-drift} and effective \emph{height-drift}.
VDL penalizes mismatches in predecessor$\rightarrow$current vertical offsets measured at three summary statistics (centroid, top band, bottom band):
{
\small
\begin{equation}
\mathcal{L}_{\mathrm{vdl}} = \frac{1}{|\mathcal{B}|} \sum_{b\in\mathcal{B}} \sum_{k \in \{\mathrm{cen}, \mathrm{top}, \mathrm{bot}\}} w_{k}\|\hat{\delta}^{k}_b-\delta^{k}_b\|_2^2,
\label{eq:vdl}
\end{equation}
}
where $\mathcal{B}$ is the set of valid bigram boundaries.
VDL uses normalized coordinates with a small coefficient (\suppsee{sec:appendix_vdl}), yielding \(\mathcal{L}_{\mathrm{bi}}=\mathcal{L}_{\mathrm{seq}}+\lambda_{\mathrm{vdl}}\mathcal{L}_{\mathrm{vdl}}\).
The total objective is \(\mathcal{L}_{\mathrm{total}}=\mathcal{L}_{\mathrm{gen}}^{(k)}+\lambda_{\mathrm{style}}\mathcal{L}_{\mathrm{style}}\), where \(\mathcal{L}_{\mathrm{style}}\) combines supervised writer- and self-supervised glyph-level contrastive losses for writer consistency and glyph-specific style.
\section{Experiments}
\label{sec:experiments}

% We evaluate \textsc{BoundInk} at both sentence and glyph levels.
% Our evaluation covers three aspects:
% (i) connectivity and spacing;
% (ii) trajectory and glyph fidelity; and
% (iii) perceptual quality, content legibility, and writer-style preservation.
% We use boundary-aware metrics, normalized trajectory distance, human preference, OCR, and writer classification.

\subsection{Experimental Setup}
\label{sec:exp_setup}

\paragraph{Boundary-Aware Metrics.}
Global trajectory similarity does not directly capture a writer's inter-character boundary behavior. We therefore introduce \emph{Connectivity and Spacing Metrics (CSM)}, comprising $\mathrm{F1}{\mathrm{Cursive}}$, CRE, KGS, and SSS. $\mathrm{F1}{\mathrm{Cursive}}$ evaluates individual connection decisions at adjacent non-space character boundaries, treating \textsc{CursiveEOC} as the positive class. CRE compares the cursive-continuation rates of generated and reference sentences. KGS evaluates horizontal gaps between adjacent non-space characters, while SSS evaluates inter-word spacing. Both use an overlap-aware symmetric log-ratio similarity that captures relative spacing errors while remaining robust to handwriting scale. Together, these metrics assess connectivity and spatial organization beyond global trajectory alignment. Complete definitions are provided in \suppref{sec:appendix_metrics}.

\paragraph{Trajectory Metrics.}

We additionally report normalized \emph{Dynamic Time Warping (DTW)} for global trajectory geometry.
For glyph-level evaluation, we report DTW under SDT-compatible English and Chinese protocols~\cite{dai2023disentangling}.
We compute DTW~\cite{berndt1994using} using Euclidean pointwise distance over $(x,y)$ trajectories and report
\[
\mathrm{DTW}_{\mathrm{norm}}
=
\mathrm{DTW}_{\mathrm{raw}}/|\mathrm{GT}|.
\]
Prediction and ground truth are independently normalized for translation and height without rotation normalization.
For methods that output trajectory deltas, we first recover absolute coordinates and then apply the same normalization.

\paragraph{Benchmark-Matched Protocols.}
Prior online handwriting generators differ in input conditions, style references, trajectory conventions, and evaluation protocols.
We therefore conduct \emph{benchmark-matched pairwise comparisons} rather than enforcing a single unified setting.
For each baseline, \textsc{BoundInk} is retrained with the corresponding split, preprocessing pipeline, and metric implementation.
We include sentence-level baselines that provide sufficient valid outputs, permit reliable character-boundary recovery, and support reproducible inference.
The resulting comparison set is compatibility-based rather than exhaustive.

For sentence-level evaluation, we use IAM-expanded for DeepWriting, BRUSH for DSD, and \textsc{CASIA-OLHWDB} 2.0--2.2 for OLHWG~\cite{ren2025decoupling,liu2011casia}.
The English protocols support evaluation of cursive continuation, inter-character spacing, and word spacing, while the Chinese protocol provides a layout-oriented setting.
For glyph-level evaluation, we use the English SDT benchmark and \textsc{CASIA-OLHWDB} 1.0--1.2 under the SDT-compatible Chinese protocol~\cite{dai2023disentangling,liu2011casia}.

\paragraph{Merged English Dataset.}
For qualitative evaluation, human evaluation, and ablation studies, we construct a merged English sentence dataset from IAM~\cite{liwicki2005iam} and BRUSH~\cite{kotani2020generating} to increase the diversity of local character transitions. Because the datasets differ in coordinate systems and trajectory density, we apply trajectory resampling and aspect-ratio-preserving height normalization with $h=1.0$. We avoid aggressive simplification to preserve writer-specific stroke details, and review and correct character segmentations following \citet{jungo2023character}. For BRUSH and DSD-style evaluation, we use a writer-disjoint split of 150 training and 20 test writers. RDP simplification~\cite{ramer1972iterative,douglas1973algorithms} is disabled during evaluation.

\paragraph{Inference and Aggregation.}

All quantitative evaluations use fully autoregressive inference without teacher forcing, with one output per test input. Metrics are computed per sample, averaged within each writer, and then macro-averaged across writers to prevent writers with more samples from dominating the results. The same coordinate recovery and normalization procedures are applied to all methods. A CSM component is reported as \texttt{--} when the corresponding protocol lacks its required boundary type.

\subsection{Quantitative Evaluation}
\label{sec:exp_quant}

\paragraph{Sentence-Level Generation.}
Table~\ref{tab:all_sentence_dtw_csm} reports results under three benchmark-matched protocols, with scores compared only within each protocol block. Under DeepWriting and DSD, \textsc{BoundInk} improves all applicable connectivity and spacing measures while reducing normalized DTW from 2.70 to 1.41 and from 1.25 to 0.99, respectively. This shows that explicit boundary modeling improves local boundary quality without sacrificing global trajectory fidelity. Under the OLHWG-compatible protocol, only KGS is defined because the evaluation set lacks cursive-positive boundaries and whitespace runs. \textsc{BoundInk} improves KGS from 0.34 to 0.40 and reduces normalized DTW from 0.17 to 0.14. For DeepWriting, outputs with segmentation errors in at least three characters are excluded from both methods following the benchmark protocol.

\begin{table}[t!]
\centering

\begin{threeparttable}
\resulttablefont
\setlength{\tabcolsep}{1.7pt}
\renewcommand{\arraystretch}{1.0}

\begin{tabular*}{\columnwidth}{@{\extracolsep{\fill}}lccccc@{}}
\toprule
\multirow{2}{*}{\textbf{Method}} &
\multicolumn{4}{c}{\textbf{CSM} ($\uparrow$)} &
\multirow{2}{*}{\textbf{DTW$_{\mathrm{norm}}$} ($\downarrow$)} \\
\cmidrule(lr){2-5}
&
\textbf{F1$_{\text{Curs}}$} &
\textbf{CRE} &
\textbf{KGS} &
\textbf{SSS} &
\\
\midrule
DeepWriting~\shortcite{aksan2018deepwriting}
    & 0.26 & 0.81 & 0.15 & 0.57 & 2.70 \\
\textbf{Ours}
    & \bfseries 0.33
    & \bfseries 0.84
    & \bfseries 0.41
    & \bfseries 0.61
    & \bfseries 1.41 \\
\midrule
DSD~\shortcite{kotani2020generating}
    & 0.07 & 0.84 & 0.39 & 0.49 & 1.25 \\
\textbf{Ours}
    & \bfseries 0.45
    & \bfseries 0.90
    & \bfseries 0.44
    & \bfseries 0.63
    & \bfseries 0.99 \\
\midrule
OLHWG~\shortcite{ren2025decoupling}
    & \textemdash\tnote{a}
    & \textemdash
    & 0.34
    & \textemdash
    & 0.17 \\
\textbf{Ours}
    & \textemdash\tnote{a}
    & \textemdash
    & \bfseries 0.40
    & \textemdash
    & \bfseries 0.14 \\
\bottomrule
\end{tabular*}
% \begin{tablenotes}[flushleft]
% \fontsize{7.0pt}{8.1pt}\selectfont
% \end{tablenotes}
\end{threeparttable}
\caption{Sentence-level CSM and normalized DTW under matched pairwise protocols. Undefined under OLHWG because the protocol contains no cursive-positive boundaries or whitespace runs.}
\label{tab:all_sentence_dtw_csm}
\end{table}

\paragraph{Glyph-Level Generation.}

Table~\ref{tab:sdt_benchmark_summary} evaluates whether the sentence-level design preserves individual glyph quality.
We compare \textsc{BoundInk} with representative character-level methods under SDT-compatible English and Chinese protocols.
We exclude \emph{Elegantly Written}~\cite{liu2024elegantly} because it enhances an observed stroke trajectory rather than generating handwriting from target text and writer references.
Despite being designed primarily for sentence-level generation, \textsc{BoundInk} obtains the lowest DTW in both protocols.
The English DTW decreases from 1.6048 with SDT to 1.3649.
The improvement in Chinese is smaller, decreasing from 0.8789 to 0.8718.
These results show that boundary-aware sentence generation preserves glyph-level trajectory fidelity.

\begin{table}[t!]
\centering

\begingroup
\small
\setlength{\tabcolsep}{2.0pt}
\renewcommand{\arraystretch}{1.0}

\begin{tabular*}{\columnwidth}{@{\extracolsep{\fill}}lcc@{}}
\toprule
\multirow{2}{*}{\textbf{Method}} &
\multicolumn{2}{c}{\textbf{DTW} ($\downarrow$)} \\
\cmidrule(lr){2-3}
& \textbf{English} & \textbf{Chinese} \\
\midrule

Drawing~\cite{zhang2017drawing}
    & 1.8519 & 1.1813 \\
DeepImitator~\cite{zhao2020deep}
    & 1.6460 & 1.0622 \\
WriteLikeYou-v2~\cite{tang2021write}
    & 1.6215 & 0.9289 \\
SDT~\cite{dai2023disentangling}
    & 1.6048 & 0.8789 \\
\textbf{Ours}
    & \bfseries 1.3649
    & \bfseries 0.8718 \\

\bottomrule
\end{tabular*}
\endgroup
\caption{Character-level DTW in SDT-compatible settings.}
\label{tab:sdt_benchmark_summary}
\end{table}

\paragraph{Component Ablation.}

We examine the contributions of Bi-SWT, gated context fusion, and the learnable character-code embedding.
Table~\ref{tab:ablation_latest_iter} presents a $2\times2$ ablation of Bi-SWT and context fusion on the merged English setting.
Removing Bi-SWT degrades all reported measures.
In particular, $\mathrm{F1}_{\mathrm{Cursive}}$ decreases from 0.49 to 0.37, SSS decreases from 0.71 to 0.40, and normalized DTW increases from 0.51 to 0.60.
These changes confirm that predecessor--current decoding contributes to both local boundary quality and global trajectory consistency.

Removing gated context fusion primarily affects cursive connectivity.
$\mathrm{F1}_{\mathrm{Cursive}}$ decreases from 0.49 to 0.35, whereas CRE, KGS, and SSS change more moderately.
This suggests that sentence context is particularly useful for deciding whether the pen should continue across a character boundary.
Local bigram modeling remains the main contributor to spacing and geometric stability.
Removing both components produces the lowest overall boundary quality.
The resulting $\mathrm{F1}_{\mathrm{Cursive}}$ of 0.22 indicates that local boundary decoding and sentence context provide complementary information.
We separately evaluate the learnable character-code embedding, denoted by CharID.
Adding CharID reduces average normalized DTW from 0.3464 to 0.3328 on the glyph-level ablation subset.
This corresponds to a 3.9\% relative improvement and supports its role in sharpening character identity.

\begin{table}[t!]
\centering

\begingroup
\small
\setlength{\tabcolsep}{1.15pt}
\renewcommand{\arraystretch}{1.0}

\begin{threeparttable}
\begin{tabular*}{\columnwidth}{@{\extracolsep{\fill}}lccccccc@{}}
\toprule
\multirow{2}{*}{\textbf{Variant}} &
\multirow{2}{*}{\shortstack{\textbf{Bi-}\\\textbf{SWT}}} &
\multirow{2}{*}{\shortstack{\textbf{Ctx.}\\\textbf{Fus.}}} &
\multicolumn{4}{c}{\textbf{CSM} ($\uparrow$)} &
\multicolumn{1}{c}{\textbf{DTW} ($\downarrow$)} \\
\cmidrule(lr){4-7}
\cmidrule(lr){8-8}
& & &
\textbf{F1$_{\text{Curs}}$} &
\textbf{CRE} &
\textbf{KGS} &
\textbf{SSS} &
\textbf{Norm} \\
\midrule

\textbf{Ours}
    & \checkmark
    & \checkmark
    & \bfseries 0.49
    & \bfseries 0.91
    & \bfseries 0.49
    & \bfseries 0.71
    & \bfseries 0.51 \\

w/o Bi-SWT
    & \textemdash
    & \checkmark
    & 0.37
    & 0.88
    & 0.44
    & 0.40
    & 0.60 \\

w/o Ctx. Fus.
    & \checkmark
    & \textemdash
    & 0.35
    & \bfseries 0.91
    & 0.48
    & 0.69
    & 0.55 \\

w/o both
    & \textemdash
    & \textemdash
    & 0.22
    & 0.85
    & 0.29
    & 0.66
    & 0.57 \\

\bottomrule
\end{tabular*}

%\vspace{0.1em}
% \begin{tablenotes}[flushleft]
% \scriptsize
% \item[] \checkmark\ denotes enabled; \textemdash\ denotes removed.
% Raw DTW is reported in the supplementary material.
%\end{tablenotes}
\end{threeparttable}
\endgroup
\caption{Ablation of Bi-SWT and gated context fusion.
\checkmark\ denotes enabled and \textemdash\ denotes removed.}
% Raw DTW is reported in the supplementary material.}
\label{tab:ablation_latest_iter}
%\vspace{-1.5mm}
\end{table}

\subsection{Perceptual and Complementary Evaluation}
\label{sec:exp_perceptual}

\paragraph{Human Preference.}

\begin{figure}[t!]
% \vspace{-3mm}
    \centering
    \includegraphics[
        width=0.92\columnwidth,
        % trim=1mm 0mm 1mm 0mm,
        % clip
    ]{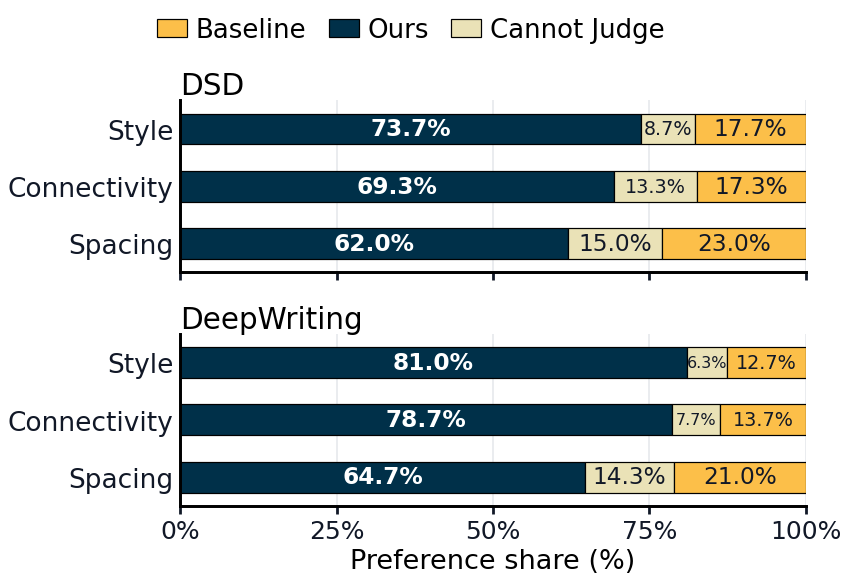}
    % \vspace{-1mm}
    \caption{Human preference study.
    Preferences for style, connectivity, and spacing against DSD and DeepWriting.}
    \label{fig:human_pref_by_metric}
    % \vspace{-1.5mm}
\end{figure}

Automated metrics do not fully capture the perceptual naturalness of writer-conditioned handwriting.
We therefore conduct a blind human preference study focused on style, inter-character connectivity, and spacing.
Thirty participants each complete 20 randomized trials, with 10 trials for each comparison protocol.
For every criterion, participants choose \textsc{BoundInk}, the corresponding baseline, or \emph{Cannot judge}.
The alternatives are presented without method identities.
\textsc{BoundInk} is preferred under both protocols.
Against DSD, it receives 615 of 789 valid criterion-wise preferences, while 111 responses are marked as \emph{Cannot judge}.
Against DeepWriting, it receives 673 of 815 valid preferences, with 85 responses marked as \emph{Cannot judge}.
Figure~\ref{fig:human_pref_by_metric} provides the criterion-wise breakdown.
The results consistently favor \textsc{BoundInk} in style, connectivity, and spacing.

\begin{figure}[t!]
    \centering
    \includegraphics[
        width=0.88\columnwidth
    ]{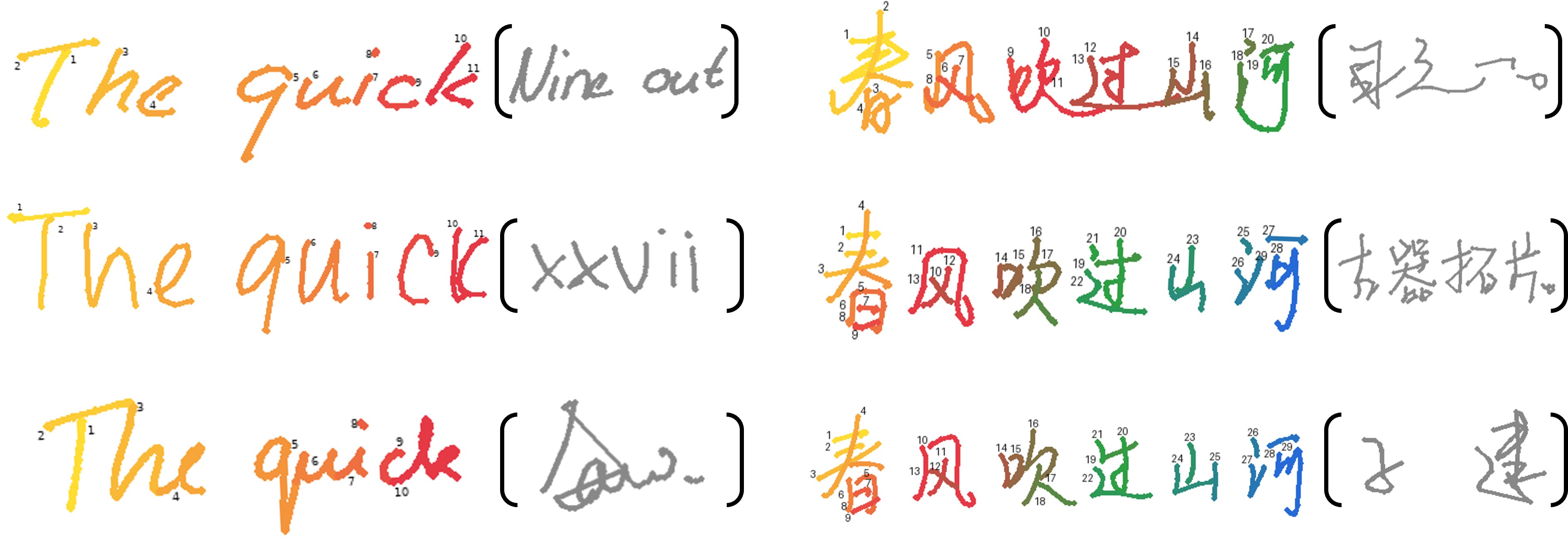}
    % \vspace{-0.5mm}
    \caption{
    Writer-style diversity under fixed content.
    Writer references produce distinct styles for the same text.
    }
    \label{fig:style_diversity_fixed_content}
    % \vspace{-1mm}
\end{figure}

\begin{figure}[t!]
    \centering
    % \small
    \setlength{\tabcolsep}{1.5pt}
    \renewcommand{\arraystretch}{0.9}

    \resizebox{\columnwidth}{!}{%
    \begin{tabular}{@{}lcc@{}}
        \toprule
        \multicolumn{3}{c}{\textbf{English}} \\
        \midrule
        GT &
        \includegraphics[
            width=0.43\columnwidth
        ]{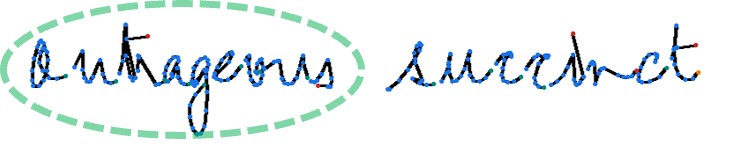} &
        \includegraphics[
            width=0.43\columnwidth
        ]{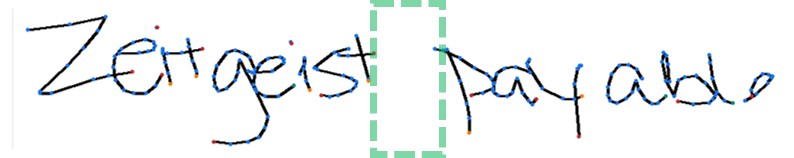} \\

        DSD~\shortcite{kotani2020generating} &
        \includegraphics[
            width=0.43\columnwidth
        ]{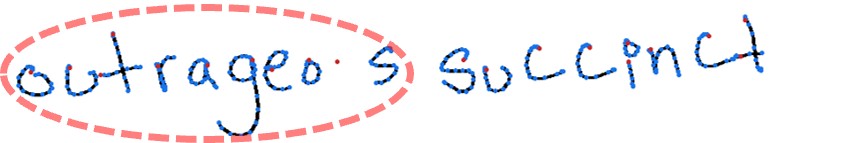} &
        \includegraphics[
            width=0.43\columnwidth
        ]{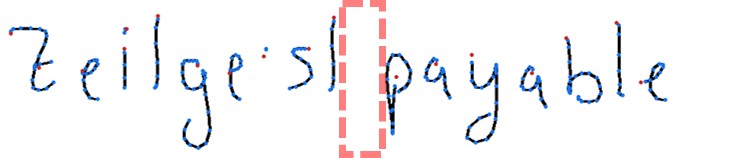} \\

        \textbf{Ours} &
        \includegraphics[
            width=0.43\columnwidth
        ]{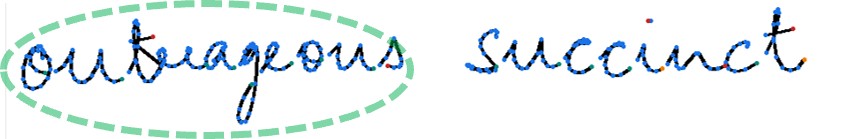} &
        \includegraphics[
            width=0.43\columnwidth
        ]{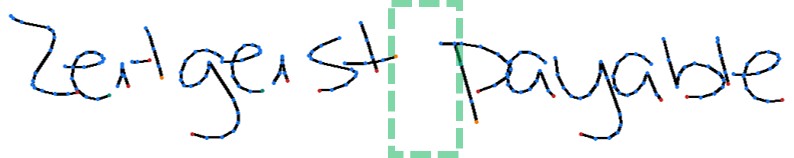} \\

        \midrule
        \multicolumn{3}{c}{\textbf{Chinese}} \\
        \midrule
        GT &
        \includegraphics[
            width=0.38\columnwidth
        ]{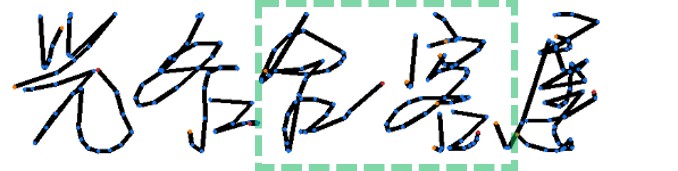} &
        \includegraphics[
            width=0.46\columnwidth
        ]{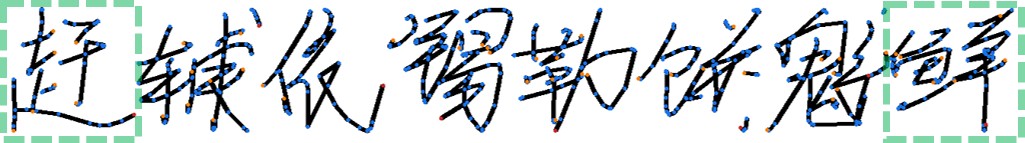} \\

        OLHWG~\shortcite{ren2025decoupling} &
        \includegraphics[
            width=0.38\columnwidth
        ]{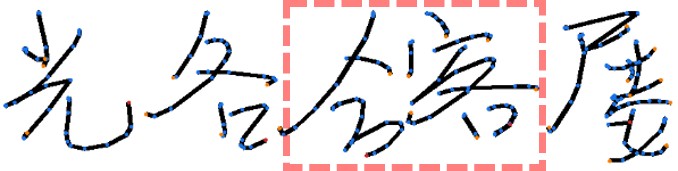} &
        \includegraphics[
            width=0.46\columnwidth
        ]{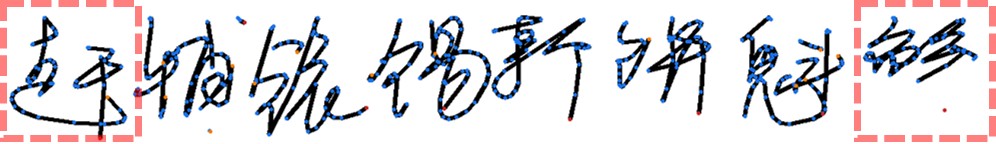} \\

        \textbf{Ours} &
        \includegraphics[
            width=0.38\columnwidth
        ]{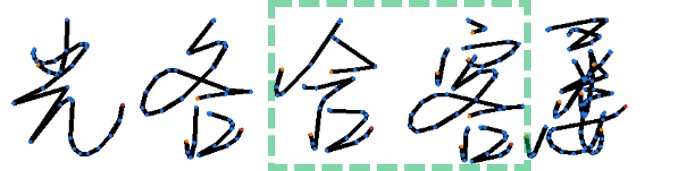} &
        \includegraphics[
            width=0.46\columnwidth
        ]{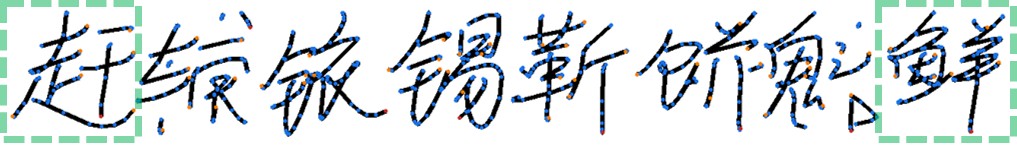} \\

        \bottomrule
    \end{tabular}%
    }

    \caption{Qualitative comparison on BRUSH and CASIA.
    \textsc{BoundInk} better preserves writer-consistent glyph structure,
    local connectivity, and spacing than the baselines.}
    \label{fig:qual_comparison}
    %\vspace{-1mm}
\end{figure}

\paragraph{Qualitative Analysis.}
Figure~\ref{fig:style_diversity_fixed_content} examines writer-style diversity under fixed textual content.
Different writer references produce distinct glyph forms and trajectory styles.
This indicates that boundary adaptation does not collapse the writer-conditioning signal.
Figure~\ref{fig:qual_comparison} compares \textsc{BoundInk} with the corresponding English and Chinese baselines.
On BRUSH, \textsc{BoundInk} better reproduces writer-specific cursive joins, local spacing, and whitespace structure.
On CASIA, cursive and whitespace boundaries are less prominent.
The main improvements instead appear in glyph structure, intra-character stroke arrangement, and spatial density.

\paragraph{Content and Writer Verification.}

\begin{figure}[t!]
\centering

% -------------------------------------------------
% (a) OCR-based verification
% -------------------------------------------------
\includegraphics[
    width=\columnwidth
]{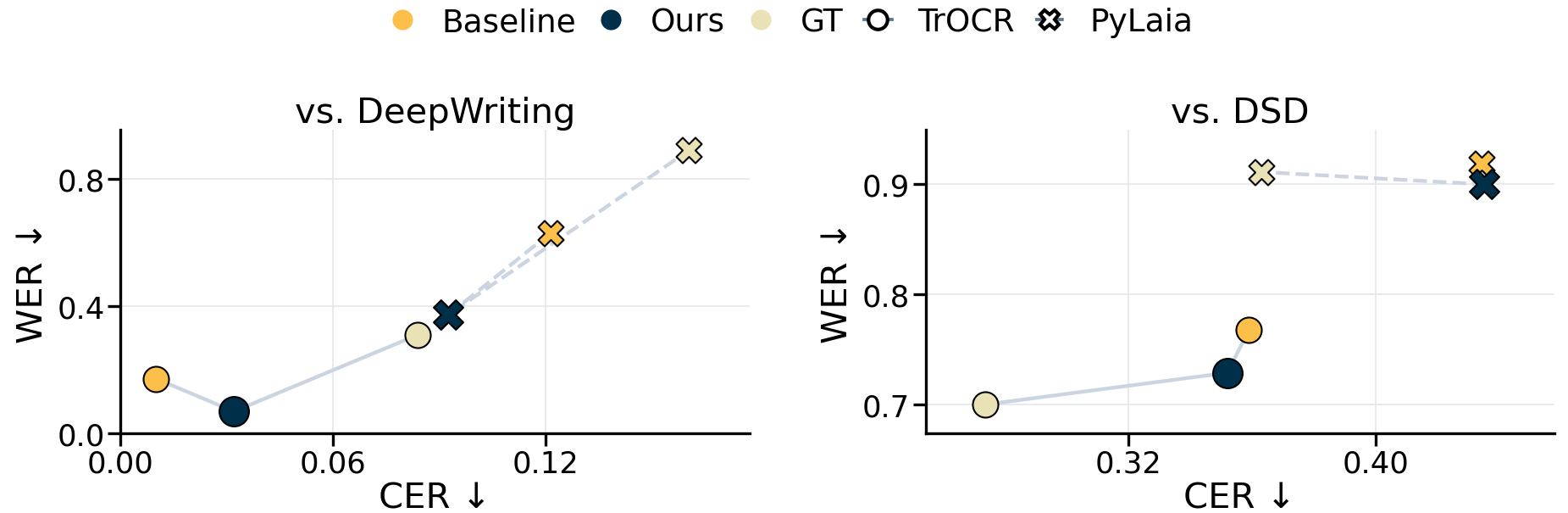}

{\small (a) OCR-based verification\par}

% -------------------------------------------------
% (b) Writer-style classification
% -------------------------------------------------
{\small
\setlength{\tabcolsep}{1.6pt}
\renewcommand{\arraystretch}{0.80}

\begin{tabular*}{\columnwidth}{
    @{\extracolsep{\fill}}
    lcc
    lcc
    @{}
}
\toprule
\multicolumn{3}{c}{\textbf{vs. DeepWriting}} &
\multicolumn{3}{c}{\textbf{vs. DSD}} \\
\cmidrule(r){1-3}
\cmidrule(l){4-6}

\textbf{Method} &
\textbf{Top-1} &
\textbf{Top-5} &
\textbf{Method} &
\textbf{Top-1} &
\textbf{Top-5} \\

\midrule

Baseline & 35.8 & 68.3 &
Baseline & 6.0 & 27.0 \\

\textbf{Ours} & \textbf{61.7} & \textbf{81.7} &
\textbf{Ours} & \textbf{69.0} & \textbf{74.0} \\

GT & 93.4 & 98.4 &
GT & 98.5 & 99.5 \\

\bottomrule
\end{tabular*}
}

{\small (b) Writer-style classification and preservation\par}

\caption{
Content and writer-style verification.
(a) OCR verification with TrOCR-large and PyLaia-IAM.
(b) Writer classification using protocol-specific classifiers trained on real handwriting.
GT denotes real handwriting.
}
\label{fig:content_style_verif}

\end{figure}

Trajectory and boundary metrics do not directly verify whether rendered outputs preserve textual content and writer identity.
We therefore perform two complementary evaluations on rendered trajectories.
OCR recognizers assess content legibility, while protocol-specific writer classifiers assess target-writer consistency.
As shown in Figure~\ref{fig:content_style_verif}, \textsc{BoundInk} remains competitive with the corresponding baselines in OCR-based recognition.
At the same time, it achieves substantially higher writer-classification accuracy under both protocols.
Against DeepWriting, Top-1 accuracy increases from 35.8 to 61.7 and Top-5 accuracy increases from 68.3 to 81.7.
Against DSD, Top-1 accuracy increases from 6.0 to 69.0 and Top-5 accuracy increases from 27.0 to 74.0.

These results show that improvements in boundary quality do not come at the expense of rendered-text legibility.
They also provide independent evidence that \textsc{BoundInk} retains stronger target-writer characteristics.

\section{Conclusion}
\label{sec:conclusion}

We presented \textsc{BoundInk}, a boundary-aware framework for sentence-level online handwriting generation.
Rather than treating inter-character transitions as incidental outcomes of long-sequence decoding, \textsc{BoundInk} explicitly models predecessor--current boundaries while preserving writer-dependent glyph appearance and incorporating sentence-level context.
Together with the proposed \emph{Connectivity and Spacing Metrics (CSM)}, our framework provides a direct means to model and evaluate cursive continuity, kerning, word spacing, and boundary alignment.
Across benchmark-matched protocols, \textsc{BoundInk} improves all applicable CSM components and normalized DTW, with consistent gains in human evaluation and ablation studies.
More broadly, this work establishes inter-character boundaries as an important modeling and evaluation unit for coherent sentence-level handwriting generation.

Future work will incorporate optional sequence references to better capture writer-specific boundary tendencies and examine generalization to broader scripts, unseen character combinations, and scarce or mismatched references. We also plan to improve long-sequence decoding efficiency and robustness to autoregressive error accumulation. Finally, standardized benchmarks are also needed to jointly assess content, style, boundary quality, reliability, and efficiency.

\clearpage

% ===============================================================
% References
% ===============================================================

\FloatBarrier
\bibliography{main}

% ===============================================================
% Appendices
% ===============================================================

\clearpage

\twocolumn[
\begin{center}
    {\LARGE\bfseries Appendices\par}
    \vspace{0.4em}
    {\large BoundInk: Boundary-Aware Online Handwriting Generation\par}
    \vspace{1.2em}
\end{center}
]

% Reset appendix figure/table/equation numbering
\setcounter{figure}{0}
\setcounter{table}{0}
\setcounter{equation}{0}

\renewcommand{\thefigure}{A\arabic{figure}}
\renewcommand{\thetable}{A\arabic{table}}
\renewcommand{\theequation}{A\arabic{equation}}

\appendix

\noindent
This \suppname{} provides additional analyses and
implementation details supporting the findings in the main paper.

% \tableofcontents
% =========================================================
% Manual Contents
% =========================================================
\noindent{\large\bfseries Contents}\par
\smallskip

\begingroup
\small

\newcommand{\manualtocsection}[3]{%
    \noindent
    \hangindent=1.6em
    \hangafter=1
    \textbf{#1}\enspace\textbf{#2}%
    \dotfill
    \pageref{#3}\par
}

\newcommand{\manualtocsubsection}[3]{%
    \noindent
    \hangindent=3.0em
    \hangafter=1
    \hspace*{1.2em}%
    #1\enspace #2%
    \dotfill
    \pageref{#3}\par
}

\manualtocsection
    {A}
    {Training Details and Curriculum Schedule}
    {sec:appendix_training_details}

\manualtocsubsection
    {A.1}
    {Curriculum and Boundary Supervision}
    {sec:appendix_curriculum_supervision}

\manualtocsubsection
    {A.2}
    {Vertical Drift Loss}
    {sec:appendix_vdl}

\manualtocsection
    {B}
    {Dataset Construction and Preprocessing}
    {sec:appendix_dataset}

\manualtocsubsection
    {B.1}
    {Merged English Dataset}
    {sec:appendix_merged_english}

\manualtocsubsection
    {B.2}
    {Chinese Protocol}
    {sec:appendix_chinese_protocol}

\manualtocsection
    {C}
    {Implementation Details of \textsc{BoundInk}}
    {sec:appendix_impl}

\manualtocsubsection
    {C.1}
    {Character Context Encoder}
    {sec:appendix_character_context_encoder}

\manualtocsubsection
    {C.2}
    {Bigram-Aware Local Decoding}
    {sec:appendix_bigram_local_decoding}

\manualtocsection
    {D}
    {Additional Experimental Analysis}
    {sec:appendix_additional_analysis}

\manualtocsubsection
    {D.1}
    {Exploratory Sequence Style Reference and Vertical Stability}
    {sec:appendix_vertical_drift_analysis}

\manualtocsection
    {E}
    {Human Evaluation Protocol}
    {sec:appendix_human_eval}

\manualtocsubsection
    {E.1}
    {Human Evaluation Design and Interface}
    {sec:appendix_human_study}

\manualtocsection
    {F}
    {Connectivity and Spacing Metrics (CSM)}
    {sec:appendix_metrics}

\manualtocsubsection
    {F.1}
    {Boundary and Cursive Metrics}
    {sec:appendix_boundary_cursive_metrics}

\manualtocsubsection
    {F.2}
    {Spacing Metrics (KGS and SSS)}
    {sec:appendix_spacing_metrics}

\manualtocsubsection
    {F.3}
    {Writer-Macro Aggregation and DTW Comparison}
    {sec:appendix_aggregation_dtw}

\endgroup
\medskip

% =========================================================
% Training Details and Curriculum Schedule
% =========================================================
\section{Training Details and Curriculum Schedule}
\label{sec:appendix_training_details}

\subsection{Curriculum and Boundary Supervision}
\label{sec:appendix_curriculum_supervision}

\begin{figure}[t!]
    \centering
    \includegraphics[width=\columnwidth]{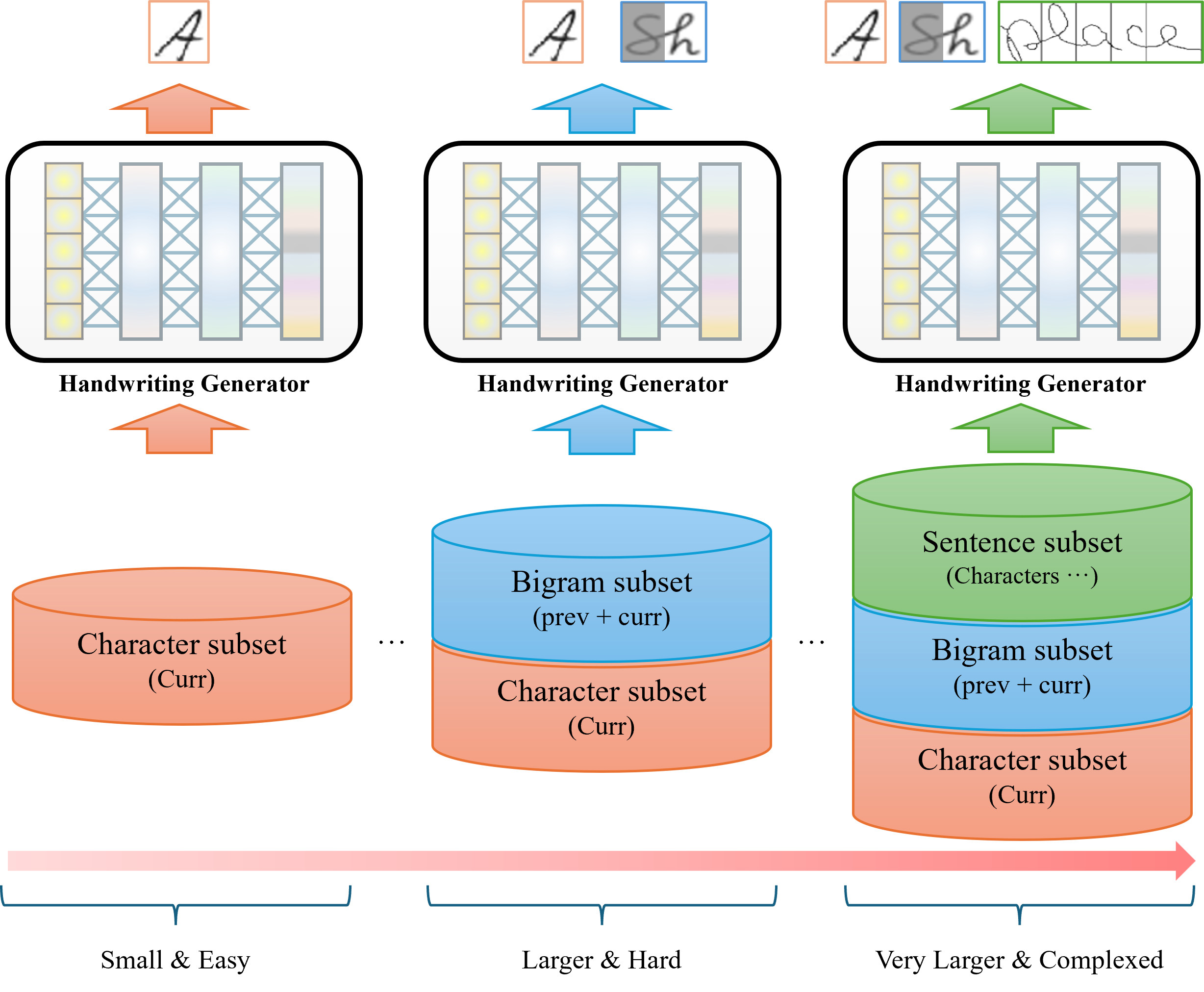}
    \caption{Stage-wise curriculum and sub-dataset composition in BoundInk.
    Stage~1 uses the character subset and learns representative glyph generation.
    Stage~2 adds the bigram subset and learns predecessor--current transitions.
    Stage~3 further adds the sentence subset and learns sentence-level generation while retaining the earlier streams.
    In the merged English setting, Stage~1 is used up to $10$k iterations, Stage~2 up to $50$k iterations, and Stage~3 for the remaining iterations.}
    \label{fig:method_stage_curriculum_learning}
\end{figure}

\begin{figure*}[t!]
    \centering
    \includegraphics[width=0.78\textwidth]{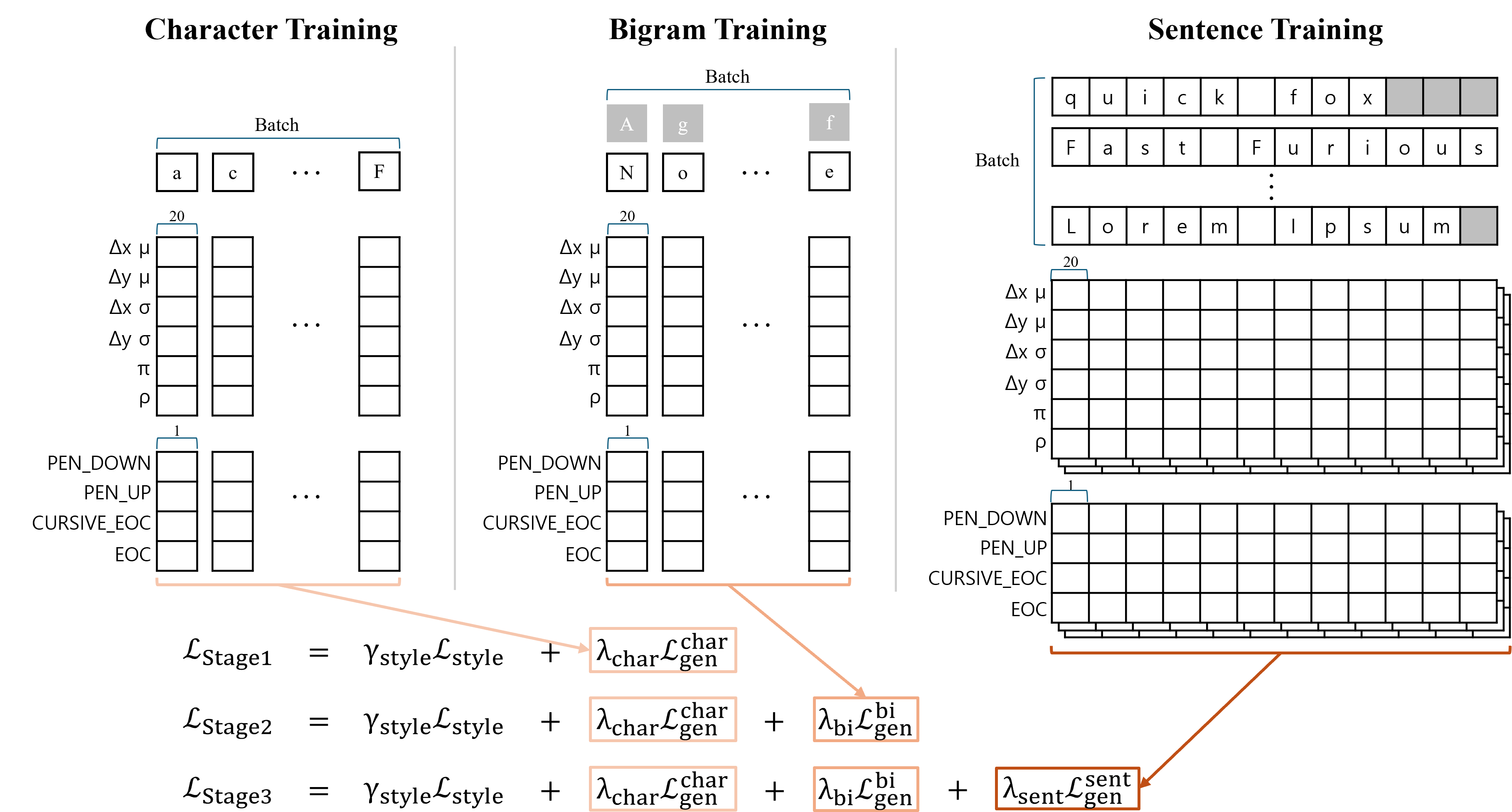}
    \caption{Cumulative supervision across stages in BoundInk.
    Stage~1 uses character-level supervision.
    Stage~2 adds bigram-level supervision together with VDL while retaining the character stream.
    Stage~3 further adds sentence-level supervision while keeping the earlier streams active.}
    \label{fig:method_visualization_of_loss}
\end{figure*}

\paragraph{Three-Stage Curriculum.}
Sentence-level online handwriting generation remains challenging under limited compositional diversity because many predecessor--current transitions are under-represented in sentence data.
To address this, \textsc{BoundInk} adopts a cumulative three-stage curriculum
(\emph{glyph} $\rightarrow$ \emph{bigram} $\rightarrow$ \emph{sentence}).
This design strengthens local boundary modeling before the model is exposed to full sentence composition and helps preserve representative glyph quality throughout training.
Figure~\ref{fig:method_stage_curriculum_learning} summarizes the stage-wise curriculum used in \textsc{BoundInk}.
Each stage is associated with a different sub-dataset and a different generation target.
As the stage increases, the training task expands from representative glyph formation to predecessor--current bigram modeling and then to full sentence generation.
Figure~\ref{fig:method_visualization_of_loss} further shows that supervision is cumulative rather than replaced across stages.

\paragraph{Training Objective.}

At stage $k$, the total objective is
\begin{equation}
\mathcal{L}_{\mathrm{total}}^{(k)}
=
\mathcal{L}_{\mathrm{gen}}^{(k)}
+
\lambda_{\mathrm{style}}^{(k)} \mathcal{L}_{\mathrm{style}},
\label{eq:appendix_total_loss}
\end{equation}
where the active generation loss is
\begin{equation}
\mathcal{L}_{\mathrm{gen}}^{(k)}
=
\lambda_{\mathrm{char}}^{(k)} \mathcal{L}_{\mathrm{char}}
+
\lambda_{\mathrm{bi}}^{(k)} \mathcal{L}_{\mathrm{bi}}
+
\lambda_{\mathrm{sent}}^{(k)} \mathcal{L}_{\mathrm{sent}}.
\label{eq:appendix_curriculum_loss}
\end{equation}
The three terms correspond to representative glyph samples, predecessor--current bigram windows, and sentence sliding windows, respectively.
For all active streams, the shared sequence-generation objective is
\begin{equation}
\mathcal{L}_{\mathrm{seq}}
=
\mathcal{L}_{\mathrm{mdn}}
+
\lambda_{\mathrm{pen}} \mathcal{L}_{\mathrm{pen}},
\label{eq:appendix_seq_loss}
\end{equation}
where $\mathcal{L}_{\mathrm{mdn}}$ is the masked GMM negative log-likelihood on valid trajectory steps and $\mathcal{L}_{\mathrm{pen}}$ is the masked cross-entropy over the four pen states
$\{\PM,\PU,\CEOC,\EOC\}$.
For the character stream, we use
\begin{equation}
\mathcal{L}_{\mathrm{char}}
=
\mathcal{L}_{\mathrm{seq}}.
\end{equation}
In the merged English setting, we use $\lambda_{\mathrm{pen}}=1.5$.
The pen-state class weights are \texttt{[1.0, 1.0, 2.0, 2.5]} for
\texttt{[PM, PU, CURSIVE\_EOC, EOC]}.

For the bigram and sentence streams, VDL is additionally applied:
\begin{gather}
\mathcal{L}_{\mathrm{bi}}
=
\mathcal{L}_{\mathrm{seq}}
+
\lambda_{\mathrm{vdl}}^{\mathrm{bi}}\mathcal{L}_{\mathrm{vdl}}, \\
\mathcal{L}_{\mathrm{sent}}
=
\mathcal{L}_{\mathrm{seq}}
+
\lambda_{\mathrm{vdl}}^{\mathrm{sent}}(t)\mathcal{L}_{\mathrm{vdl}}.
\label{eq:appendix_bigram_sentence_with_vdl}
\end{gather}

The style objective is
\begin{equation}
\mathcal{L}_{\mathrm{style}}
=
\mathcal{L}_{\mathrm{SupCon}}^{\mathrm{writer}}
+
\mathcal{L}_{\mathrm{SupCon}}^{\mathrm{glyph}}.
\label{eq:appendix_style_loss}
\end{equation}

Architecture-specific details such as the context encoder, Bi-SWT decoder, and GMM output head are described separately in Section~\ref{sec:appendix_impl}.

\paragraph{Optimization and Stage Schedule.}

All experiments in the reported merged English setting were conducted on a single NVIDIA H100 GPU.
Under this setup, full training required approximately 10 days.

In the merged English setting, Stage~1 is used for the first $10$k iterations, Stage~2 from $10$k to $50$k iterations, and Stage~3 for the remaining iterations.
The curriculum is cumulative: Stage~1 activates only the character stream, Stage~2 activates both the character and bigram streams, and Stage~3 activates character, bigram, and sentence streams together.
Stage transitions use an additional \texttt{1k}-iteration warmup to reduce instability caused by abrupt changes in the effective data distribution.
Table~\ref{tab:appendix_curriculum_schedule} summarizes the stage-wise curriculum, active streams, stage-level coefficients, and nominal minibatch ratios used in the merged English configuration.
Stage~2 emphasizes bigram generation while retaining a reduced character stream, and Stage~3 shifts the main emphasis to sentence generation while preserving lower-level supervision.
The nominal Stage~3 minibatch ratio is \texttt{1:1:4}, reflecting one character batch, one bigram batch, and sentence batches accumulated over four steps.

\begin{table}[t!]
\centering

\small
\setlength{\tabcolsep}{1.2pt}
\renewcommand{\arraystretch}{1.0}

\begin{tabular*}{\columnwidth}{
@{\extracolsep{\fill}}cccccccc@{}
}
\toprule
\textbf{Stg.} &
\textbf{Iter.} &
\textbf{Char} &
\textbf{Bi} &
\textbf{Sent} &
$\lambda_{\text{sty}}$ &
$\lambda_{\text{vdl}}$ &
\textbf{Ratio} \\
\midrule
1 & $[0,10\mathrm{k})$
  & 1 & 0 & 0 & 1.0 & 0
  & 1:0:0 \\
2 & $[10\mathrm{k},50\mathrm{k})$
  & 1 & 1 & 0 & 0.3 & 0.01
  & 1:1:0 \\
3 & $[50\mathrm{k},10^6]$
  & 1 & 1 & 1 & 0.2
  & Eq.~\eqref{eq:appendix_vdl_schedule}
  & 1:1:4 \\
\bottomrule
\end{tabular*}

\caption{Stage-wise curriculum schedule in the merged English sentence setting.
Char, Bi, and Sent indicate active training streams; step-specific generation-loss weights are given below.}
\label{tab:appendix_curriculum_schedule}
\end{table}

\paragraph{Implementation Mapping.}
The merged English experiments use the following step-specific loss weights.
For Stage~1, we use:
\begin{itemize}
    \item \texttt{CHAR\_STEP\_STYLE\_LOSS\_WEIGHT=1.0}
    \item \texttt{CHAR\_STEP\_CHAR\_GEN\_LOSS\_WEIGHT=1.0}
\end{itemize}
For Stage~2, we use:
\begin{itemize}
    \item \texttt{BIGRAM\_STEP\_STYLE\_LOSS\_WEIGHT=0.3}
    \item \texttt{BIGRAM\_STEP\_CHAR\_GEN\_LOSS\_WEIGHT=0.3}
    \item \texttt{BIGRAM\_STEP\_BIGRAM\_GEN\_LOSS\_WEIGHT=1.0}
\end{itemize}
For Stage~3, we use:
\begin{itemize}
    \item \texttt{BIGRAM\_STEP\_BIGRAM\_GEN\_}\\\texttt{LOSS\_WEIGHT=1.0}
    \item \texttt{STAGE3\_BIGRAM\_STEP\_}\\\texttt{BIGRAM\_GEN\_LOSS\_WEIGHT=0.5}
    \item \texttt{SENT\_STEP\_STYLE\_LOSS\_WEIGHT=0.2}
    \item \texttt{SENT\_STEP\_CHAR\_GEN\_LOSS\_WEIGHT=0.3}
    \item \texttt{STAGE3\_USE\_BIGRAM=true}
\end{itemize}
Accordingly, Stage~1 focuses on representative glyph formation, Stage~2 introduces predecessor--current boundary modeling, and Stage~3 adds sentence-level generation while retaining the lower-level streams.

Table~\ref{tab:appendix_training_hparams} summarizes the main optimization hyperparameters used in the merged English sentence setting.

\begin{table}[t]
\centering
\small
\setlength{\tabcolsep}{4pt}
\renewcommand{\arraystretch}{1.10}
\begin{tabular}{@{} l >{\raggedright\arraybackslash}p{4.2cm} @{}}
\toprule
\textbf{Field} & \textbf{Setting} \\
\midrule
Optimizer & AdamW (\texttt{fused=True}) \\
Base learning rate & \texttt{8e-5} \\
Max iterations & \texttt{1{,}000{,}000} \\
Initial warmup & \texttt{30k} iterations \\
Stage-transition warmup & \texttt{1k} iterations \\
Mixed precision & \texttt{USE\_AMP=false} \\
Gradient clipping & global norm \texttt{1.0} for all stages \\
Batch size & \texttt{CHAR=64}, \texttt{BIGRAM=64}, \texttt{SENT=16} \\
Gradient accumulation & Stage~1 \texttt{1}, Stage~2 \texttt{1}, Stage~3-char \texttt{1}, Stage~3-sent \texttt{4} \\
Contrastive temperature & Stage~2/3: \texttt{0.07} \\
\bottomrule
\end{tabular}

\caption{Optimization and training hyperparameters used in the merged English sentence setting.}
\label{tab:appendix_training_hparams}
\end{table}

\subsection{Vertical Drift Loss}
\label{sec:appendix_vdl}

\textsc{BoundInk} predicts trajectories in delta coordinates.
This reduces per-step prediction difficulty, but sentence generation remains vulnerable to accumulated predecessor errors, which can induce vertical misalignment across character boundaries.
To reduce such drift, we use a lightweight Vertical Drift Loss (VDL) on valid adjacent character boundaries.
In the merged English setting, VDL is introduced in Stage~2 through the bigram stream and remains active in Stage~3 through the sentence stream.
Let the predicted delta trajectory be converted to absolute coordinates by cumulative summation.
For each character, we extract three vertical summary statistics from the absolute trajectory, namely the \emph{centroid}, \emph{top band}, and \emph{bottom band}.
For a valid adjacent boundary $b=(s-1,s)\in\mathcal{B}$, we define the previous-to-current vertical offsets as
\begin{align}
\delta^{\mathrm{cen}}_b &= \mathrm{cen}(s) - \mathrm{cen}(s-1), \\
\delta^{\mathrm{top}}_b &= \mathrm{top}(s) - \mathrm{top}(s-1), \\
\delta^{\mathrm{bot}}_b &= \mathrm{bot}(s) - \mathrm{bot}(s-1),
\end{align}
and analogously $\hat{\delta}^{\mathrm{cen}}_b$, $\hat{\delta}^{\mathrm{top}}_b$, and $\hat{\delta}^{\mathrm{bot}}_b$ for the prediction.
The VDL objective is then
\begin{equation}
\begin{split}
\mathcal{L}_{\mathrm{vdl}} &= \frac{1}{|\mathcal{B}|} \sum_{b\in\mathcal{B}} \Big( w_{\mathrm{cen}} \|\hat{\delta}^{\mathrm{cen}}_b-\delta^{\mathrm{cen}}_b\|_2^2 \\
&\quad + w_{\mathrm{top}} \|\hat{\delta}^{\mathrm{top}}_b-\delta^{\mathrm{top}}_b\|_2^2 + w_{\mathrm{bot}} \|\hat{\delta}^{\mathrm{bot}}_b-\delta^{\mathrm{bot}}_b\|_2^2 \Big).
\end{split}
\label{eq:appendix_vdl}
\end{equation}

In the current merged English configuration, we use
\begin{equation}
w_{\mathrm{cen}}=2.0, \quad
w_{\mathrm{top}}=1.0, \quad
w_{\mathrm{bot}}=1.0.
\end{equation}
The bigram stream uses a fixed VDL coefficient from Stage~2 onward:
\begin{equation}
\lambda_{\mathrm{vdl}}^{\mathrm{bi}} = 0.01,
\qquad
t \ge 10\mathrm{k}.
\label{eq:appendix_vdl_bigram}
\end{equation}
After Stage~3 begins, sentence-level VDL is warmed up separately:
\begin{equation}
\lambda_{\mathrm{vdl}}^{\mathrm{sent}}(t)=
\begin{cases}
0, & 50\mathrm{k} \le t < 60\mathrm{k},\\[2pt]
0.02 \min\!\left(1,\frac{t-60\mathrm{k}}{20\mathrm{k}}\right), & 60\mathrm{k} \le t < 80\mathrm{k},\\[2pt]
0.02, & t \ge 80\mathrm{k}.
\end{cases}
\label{eq:appendix_vdl_schedule}
\end{equation}
Thus, bigram-level VDL is active during Stage 2. In Stage 3, the separate bigram VDL term is disabled, and VDL is instead applied through the sentence stream according to the sentence-level warmup schedule.

\begin{figure}[t!]
    \centering
    \includegraphics[
        width=0.72\columnwidth
    ]{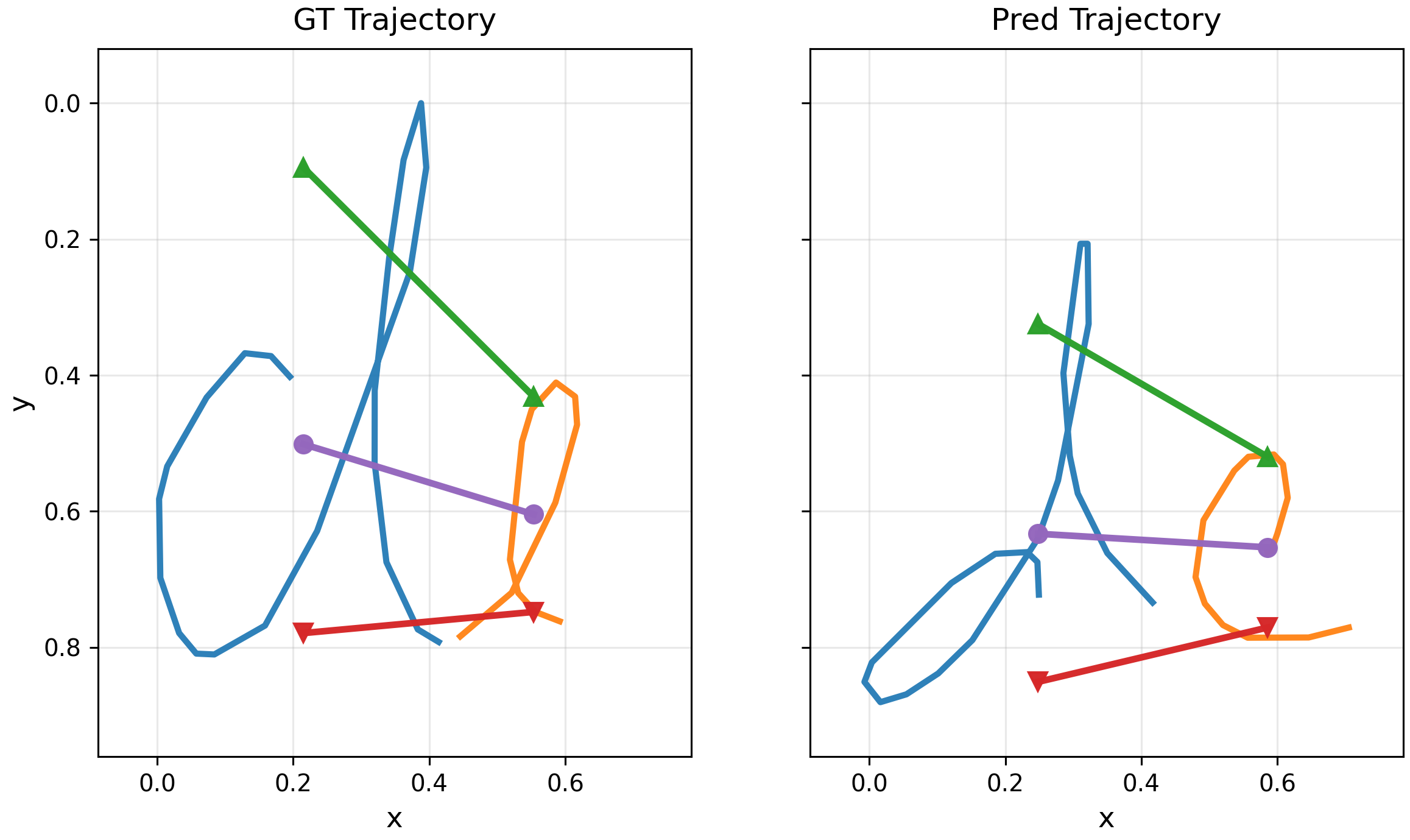}
    \caption{Bigram trajectories for ``de''.
    Ground-truth and predicted trajectories are shown with the top, centroid, and bottom statistics used by VDL.}
    \label{fig:vdl_bigram_traj_overlay}
\end{figure}

\begin{figure}[t!]
    \centering
    \includegraphics[
        width=0.55\columnwidth
    ]{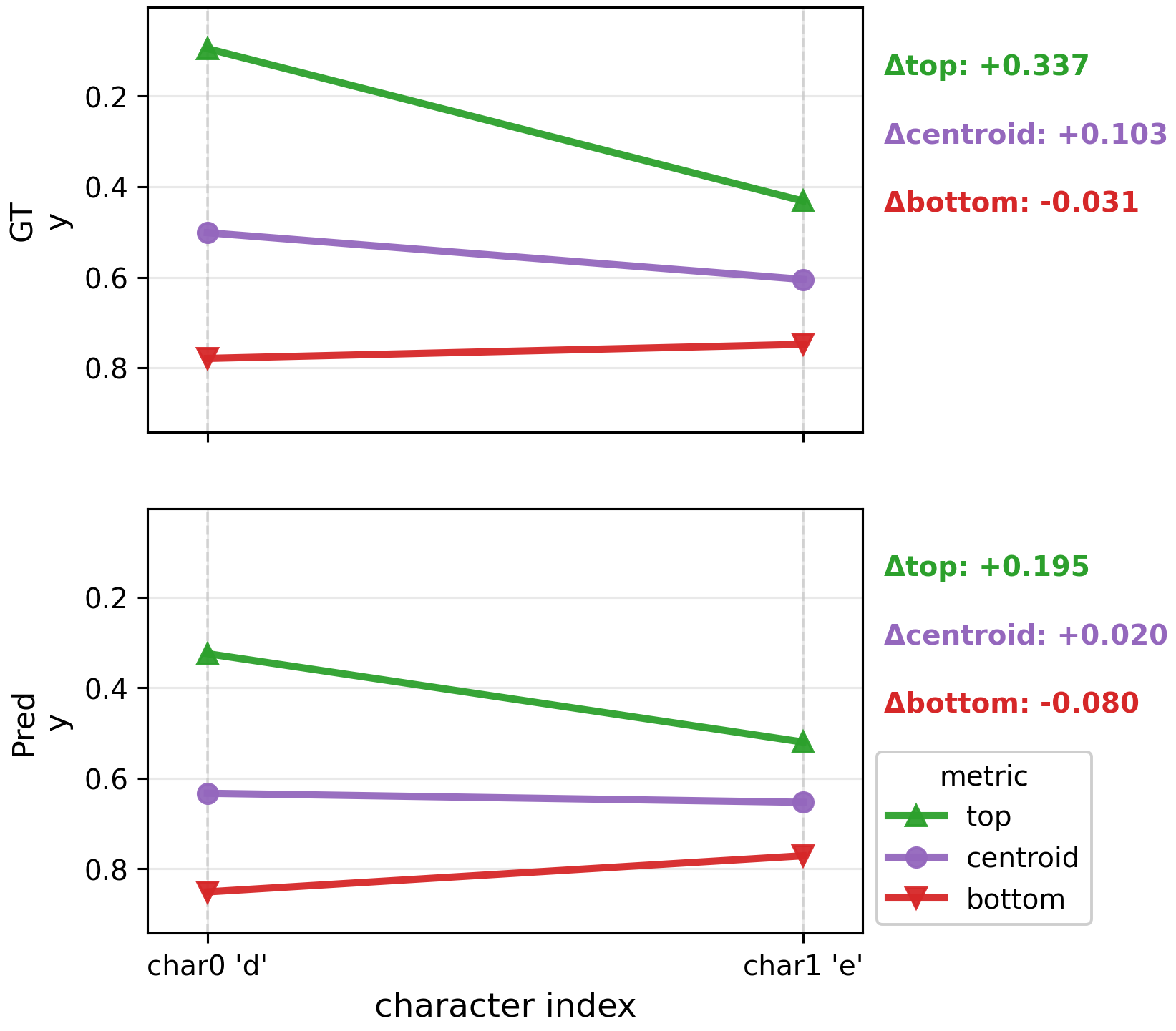}
    \caption{Boundary offsets for the same bigram ``de''.
    The previous-to-current offsets of the top, centroid, and bottom statistics are used by VDL to regularize relative vertical alignment.}
    \label{fig:vdl_bigram_char_stats}
\end{figure}

\paragraph{Numerical Example of VDL.}
Figures~\ref{fig:vdl_bigram_traj_overlay} and
\ref{fig:vdl_bigram_char_stats} provide a concrete example on the bigram boundary $b=(d,e)$.
In this case, $|\mathcal{B}|=1$.
Using the measured values shown in the figure, the ground-truth and predicted vertical offsets are
\begin{align}
\delta_b^{\mathrm{top}} &= 0.336806, & \hat{\delta}_b^{\mathrm{top}} &= 0.195319,\\
\delta_b^{\mathrm{cen}} &= 0.103238, & \hat{\delta}_b^{\mathrm{cen}} &= 0.019994,\\
\delta_b^{\mathrm{bot}} &= -0.031129, & \hat{\delta}_b^{\mathrm{bot}} &= -0.079883.
\end{align}
The corresponding squared errors are
\begin{align}
\ell_{\mathrm{top}} &= (\hat{\delta}_b^{\mathrm{top}}-\delta_b^{\mathrm{top}})^2 = 0.020018,\\
\ell_{\mathrm{cen}} &= (\hat{\delta}_b^{\mathrm{cen}}-\delta_b^{\mathrm{cen}})^2 = 0.006930,\\
\ell_{\mathrm{bot}} &= (\hat{\delta}_b^{\mathrm{bot}}-\delta_b^{\mathrm{bot}})^2 = 0.002377.
\end{align}
Using $w_{\mathrm{cen}}=2.0$ and $w_{\mathrm{top}}=w_{\mathrm{bot}}=1.0$, we obtain
\begin{align}
\mathcal{L}_{\mathrm{vdl}} &= \frac{1}{|\mathcal{B}|} \sum_{b\in\mathcal{B}} \left( 1.0\,\ell_{\mathrm{top}} + 2.0\,\ell_{\mathrm{cen}} + 1.0\,\ell_{\mathrm{bot}} \right) \\
&= 1.0(0.020018) + 2.0(0.006930) + 1.0(0.002377) \\
&= 0.036255.
\end{align}
This example shows that VDL is determined by relative boundary-wise vertical mismatch.
In practice, this term is active in Stage~2 at the bigram level and is later strengthened in Stage~3 through the sentence-level warmup schedule in Eq.~\eqref{eq:appendix_vdl_schedule}.

% =========================================================
% Dataset Construction and Preprocessing
% =========================================================
\section{Dataset Construction and Preprocessing}
\label{sec:appendix_dataset}

\subsection{Merged English Dataset}
\label{sec:appendix_merged_english}

\paragraph{IAM--BRUSH Harmonization.}
For qualitative examples, human evaluation, and ablation, we construct a merged English sentence dataset from \textsc{IAM-OnDB}~\cite{liwicki2005iam} and \textsc{BRUSH}~\cite{kotani2020generating}.
The goal of this merged setting is to increase the diversity of predecessor--current character transitions available to the Bi-SWT decoder.
This is particularly useful for modeling boundary-local phenomena such as cursive joins, kerning, and local spacing patterns.
Although both datasets provide online English handwriting trajectories, they differ in coordinate conventions, point-density distributions, and sentence-level geometric bias.
A naive merge can therefore introduce domain mismatch, distort trajectory statistics, and destabilize boundary behavior across datasets.
To reduce this mismatch, we apply a harmonization pipeline before combining the data.

\begin{figure}[t!]
    \centering
    \includegraphics[width=\columnwidth]{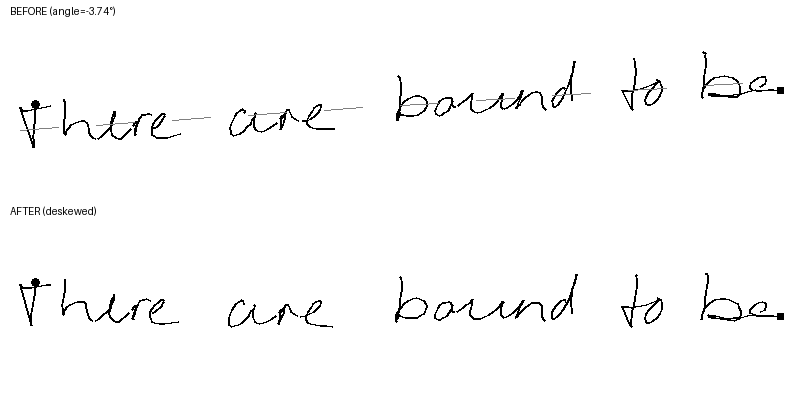}
    \caption{Sentence-level deskewing on \textsc{IAM-OnDB}~\cite{liwicki2005iam}.
    We estimate the global writing angle by least-squares line fitting and rotate the trajectory to reduce sentence-level skew while preserving the original shape.}
    \label{fig:preprocessing_deskew_sentence}
\end{figure}

\paragraph{Deskewing and Segmentation Review.}
We observed that \textsc{IAM-OnDB}~\cite{liwicki2005iam} sentences often exhibit sentence-level skew.
For this reason, we apply a lightweight deskewing procedure to IAM samples before merging.
Given the sentence trajectory, we fit a line
\begin{equation}
y = mx + b,
\end{equation}
by least-squares estimation and compute the sentence angle as
\begin{equation}
\theta = \arctan(m).
\end{equation}
The trajectory is then rotated by $-\theta$ to reduce the estimated sentence-level skew.
To avoid unnecessary or unstable correction, we skip deskewing when the estimated angle is either too small to be meaningful or too large to be reliable.
Figure~\ref{fig:preprocessing_deskew_sentence} shows a representative example.

This preprocessing is applied only to the IAM~\cite{liwicki2005iam} side because BRUSH does not show the same systematic skew pattern in our merged-data preparation.

Because the merged setting is used for connectivity-sensitive sentence generation, we additionally review character segmentation and boundary consistency on the IAM~\cite{liwicki2005iam} side.
In particular, we follow a segmentation-review procedure inspired by \emph{Character Queries}~\cite{jungo2023character} to identify and correct obvious segmentation inconsistencies before building the merged set.
This step is important because boundary-aware supervision in \textsc{BoundInk}, including \CEOC-based connectivity modeling, depends on reliable character boundaries.
The merged English dataset is used for qualitative examples, human evaluation design support, and the ablation setting reported in the paper.
Its role is to provide richer local transition coverage for sentence generation.
However, this merged setting is not used for benchmark-matched quantitative comparison.
For pairwise benchmark comparisons, \textsc{BoundInk} is retrained under each benchmark-specific protocol rather than evaluated on the merged set.
This keeps the comparisons compatible with the conventions of the corresponding baseline setting.

\paragraph{Trajectory Resampling and RDP.}

\textsc{IAM-OnDB}~\cite{liwicki2005iam} and \textsc{BRUSH}~\cite{kotani2020generating} differ not only in coordinate convention but also in the number of trajectory points per character.
To reduce this mismatch, we primarily use arc-length-based resampling so that the number of points can be adjusted while preserving the overall stroke shape as much as possible.
When further reduction is needed, we apply only a weak Ramer--Douglas--Peucker (RDP)~\cite{ramer1972iterative,douglas1973algorithms} simplification.

\begin{figure}[t!]
    \centering
    \includegraphics[
        width=0.72\columnwidth
    ]{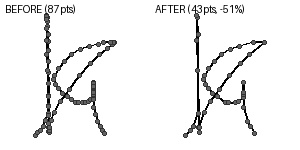}
    \caption{Point-density harmonization using resampling and weak RDP.
    The number of points is substantially reduced while preserving the overall trajectory shape.}
    \label{fig:preprocessing_resample_rdp_1}
\end{figure}

\begin{figure}[t!]
    \centering
    \includegraphics[
        width=0.72\columnwidth
    ]{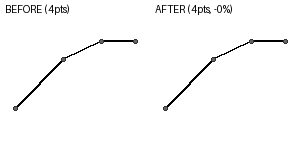}
    \caption{Example in which resampling and RDP are skipped.
    The original trajectory is retained when point reduction would provide little benefit or distort the stroke shape.}
    \label{fig:preprocessing_resample_rdp_2}
\end{figure}

Figures~\ref{fig:preprocessing_resample_rdp_1} and
\ref{fig:preprocessing_resample_rdp_2} show representative examples.

We intentionally prefer resampling over aggressive simplification because excessive trajectory reduction can damage local shape cues and degrade boundary realism.
Importantly, the start and end points of every stroke are preserved during preprocessing.
This prevents corruption of explicit pen-state events such as \PU, \EOC, and \CEOC, which are later used for connectivity-aware supervision and evaluation.

\paragraph{Height Normalization.}

After deskewing and segmentation review, all sentence trajectories are normalized into a shared coordinate system.
Following the normalization style adopted in the OLHWG~\cite{ren2025decoupling}-compatible setting, we normalize each sentence to have height $1.0$ while preserving aspect ratio.
This harmonizes the overall spatial scale across datasets without distorting writer-specific horizontal spacing patterns.
The same normalization principle is also applied to downstream sentence-level preprocessing in order to keep trajectory scale comparable across samples and across datasets.
In particular, scale harmonization is performed after geometric cleanup so that skew correction and point-density processing do not interact with inconsistent raw coordinate ranges.

\subsection{Chinese Protocol}
\label{sec:appendix_chinese_protocol}

For the Chinese experiments, we follow the same comparison protocol used in the OLHWG~\cite{ren2025decoupling} study.
Specifically, we use the OLHWD~\cite{liu2011casia} data provided by that study, which is based on \textsc{CASIA-OLHWDB}~2.0--2.2.
To keep the comparison aligned with OLHWG, the train/test split is defined by writer ID, using writers \texttt{0--1019} for training and writers \texttt{1020--1218} for evaluation.
To minimize protocol differences other than model architecture, the preprocessing outputs are converted into the unified \textsc{BoundInk} sub-dataset format used for training, namely character-level, sentence-level, and style-reference subsets.
This allows the Chinese experiments to follow the same training pipeline as the English setting while preserving the writer split and sentence protocol adopted in the OLHWG~\cite{ren2025decoupling} benchmark.

We apply strict sentence-segmentation validation during preprocessing.
If the sum of character durations does not match the total trajectory length, the sentence is discarded.
Likewise, if a character boundary implied by the duration annotations does not align with a stroke end marker within a tolerance of $\pm 3$ points, the sentence is discarded.
These checks reduce segmentation noise before boundary-aware training and evaluation.

The coordinate transformation and normalization order is also fixed explicitly.
First, the original relative coordinates are converted into absolute coordinates by cumulative summation.
Next, deskewing is applied at the sentence level.
Finally, the full sentence trajectory is normalized to height $1.0$.
This ordering stabilizes global sentence geometry before local character windows are extracted.
Long Chinese character trajectories are additionally stabilized by point-count control.
When the number of points in a character exceeds a predefined threshold (default: $160$), the character trajectory is resampled before being written into the final training subset.
This prevents unusually long trajectories from dominating sequence length statistics while preserving the overall stroke structure needed for online handwriting generation.

% =========================================================
% Implementation Details of BoundInk
% =========================================================
\section{Implementation Details of \textsc{BoundInk}}
\label{sec:appendix_impl}

In the reported model, we use \texttt{google/canine-c}~\cite{clark2022canine} as the text backbone and a ResNet18~\cite{he2016deep}-based style identifier.
We choose CANINE primarily for training efficiency while retaining character-level text encoding capability, although the framework is not restricted to CANINE and could be paired with other character-aware text encoders.
For style encoding, we adopt ResNet18 by following the stylization strategy of SDT~\cite{dai2023disentangling}, since the style input consists of binary single-channel handwriting images and a lightweight convolutional backbone is more suitable for feature extraction than heavier vision backbones.
In our implementation, the CANINE~\cite{clark2022canine} backbone is frozen during training, whereas the ResNet18~\cite{he2016deep} style encoder is optimized jointly with the rest of the model.
This section describes the main architectural components of \textsc{BoundInk}.
Optimization settings, stage-wise schedules, and other training hyperparameters are reported separately in Section~\ref{sec:appendix_training_details}.

\subsection{Character Context Encoder}
\label{sec:appendix_character_context_encoder}

The Character Context Encoder provides two complementary text-side signals for sentence-level generation.
It produces a Character-Identity Embedding that specifies which character to draw and a position-dependent Context Memory that modulates how it should appear and connect in sentence context.
This role split allows \textsc{BoundInk} to preserve character correctness while adapting character realization to local sentence context.

% \begin{figure}[t!]
%     \centering
%     \includegraphics[width=\columnwidth]{figure/appendix_context_memory_example.jpg}
%     \caption{Single-sentence probe example.}
%     \label{fig:supp_probe_example}
% \end{figure}

For a Unicode character $u$, we encode it in isolation and obtain a deterministic identity embedding
\begin{equation}
\mathbf{e}^{\mathrm{id}}(u) = f_{\mathrm{id}}\!\left(f_{\mathrm{text}}([u])\right) \in \mathbb{R}^{D}.
\label{eq:appendix_eid}
\end{equation}
Encoding the character alone avoids contextual leakage and makes the identity pathway act as a stable anchor for content specification.
In practice, the identity embedding is cached for unique codepoints within a mini-batch for efficient reuse.
To sharpen discrete identity signals, \textsc{BoundInk} uses the augmented identity embedding
\begin{equation}
\tilde{\mathbf{e}}^{\mathrm{id}}(u)
=
\mathbf{e}^{\mathrm{id}}(u)
+
\alpha \mathbf{P}\phi(u),
\label{eq:appendix_augmented_eid}
\end{equation}
where $\phi(u)$ is a learnable code embedding, $\mathbf{P}$ is a projection matrix, and $\alpha$ is a learnable scalar.
This improves identity separability while keeping identity specification separate from sentence-dependent modulation.

For a sentence-level character sequence $\mathbf{u}=(u_1,\ldots,u_S)$, we compute sentence-aware context features as
\begin{gather}
\mathbf{H}^{\mathrm{text}} = \mathbf{W}_{h} f_{\mathrm{text}}(\mathbf{u}), \\
\mathbf{M}^{\mathrm{ctx}} = f_{\mathrm{ctx}}\!\left(\mathrm{TransEnc}_{\mathrm{ctx}}(\mathbf{H}^{\mathrm{text}})\right).
\label{eq:appendix_context_memory}
\end{gather}
The vector $\mathbf{m}^{\mathrm{ctx}}_s$ varies with neighboring characters and sentence position.
This makes it suitable for modeling context-dependent realization and inter-character connectivity such as kerning and cursive joins.

\begin{figure}[t!]
    \centering
    \includegraphics[width=\columnwidth]{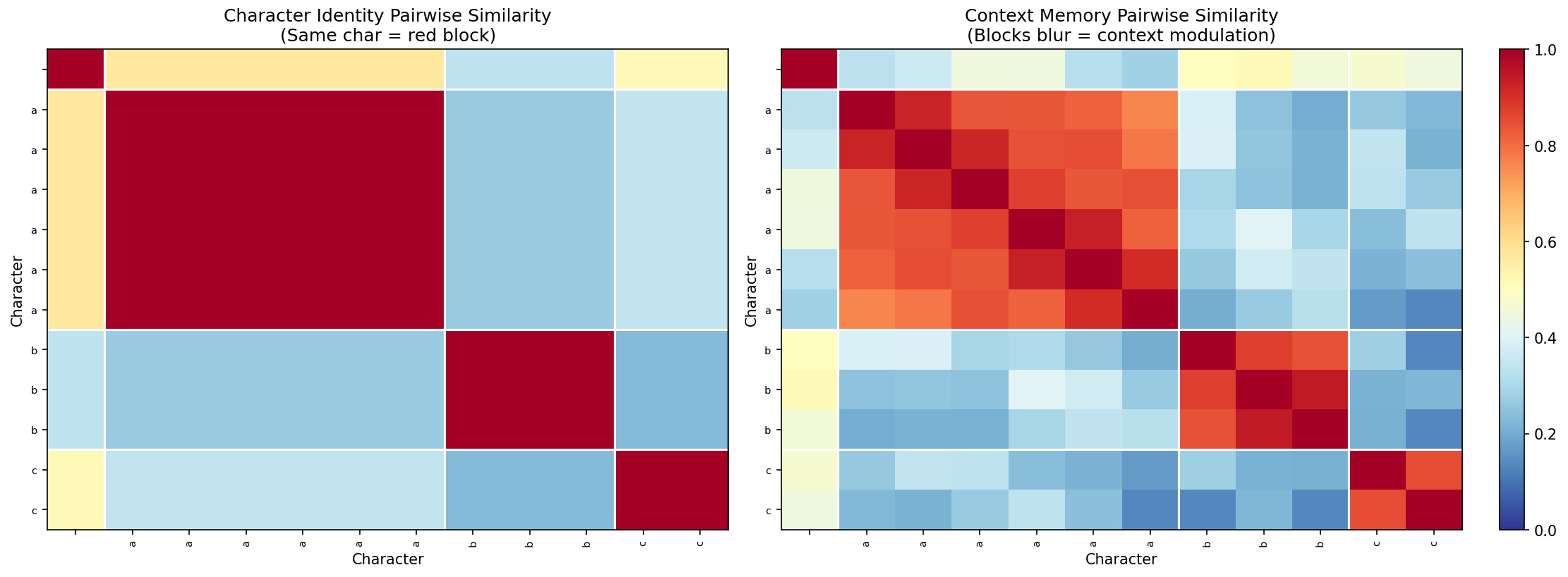}
    \caption{Pairwise cosine-similarity matrices for
    (a) Character-Identity Embedding and (b) Context Memory on the probe
    sentence ``abaca cababa.''}
    \label{fig:supp_probe_similarity}
\end{figure}

\begin{figure}[t!]
    \centering
    \includegraphics[width=\columnwidth]{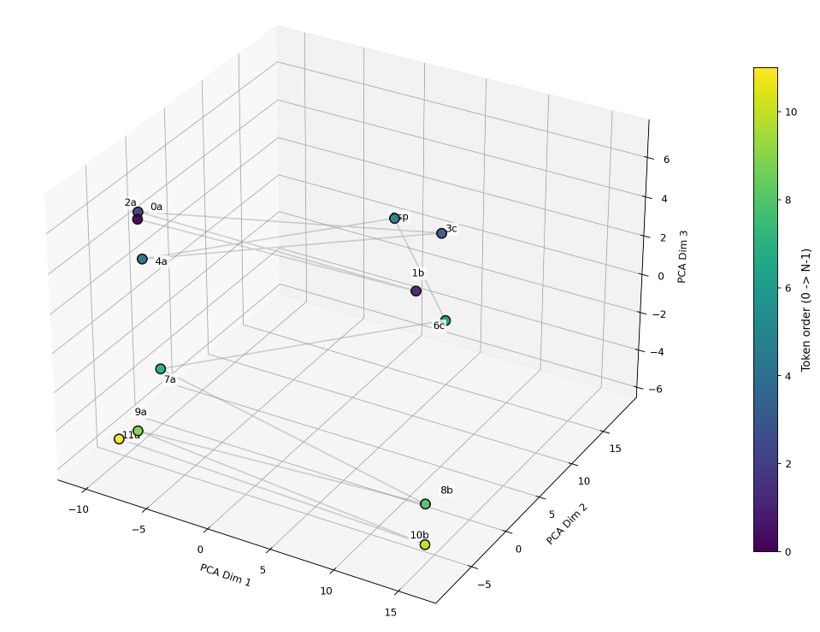}
    \caption{3D PCA projection of Context Memory tokens.}
    \label{fig:supp_probe_pca}
\end{figure}

To make this role split concrete, we analyze the two outputs of the Character Context Encoder on a single-sentence probe, ``abaca cababa''.
This probe contains $12$ tokens in total, namely six \textit{a}'s, three \textit{b}'s, two \textit{c}'s, and one space.
Figure~\ref{fig:supp_probe_similarity} and \ref{fig:supp_probe_pca} visualize the resulting representations.

Since the Character-Identity Embedding is computed from isolated-character inputs, it is expected to remain deterministic for each unique codepoint.
By contrast, Context Memory is computed from the full sentence and can therefore vary with local position and neighboring characters.

\paragraph{Character-Identity Embedding.}
Panel~(a) of Fig.~\ref{fig:supp_probe_similarity} shows sharp same-character similarity blocks for the Character-Identity Embedding, indicating that repeated occurrences of the same character collapse to a stable character-level identity.
Quantitatively, the mean same-character similarity is $1.000$, whereas the mean different-character similarity is $0.337$, yielding a separation gap of $0.663$.
By contrast, Panel~(b) shows more diffuse same-character blocks for Context Memory. Repeated occurrences of the same character remain related but no longer collapse to a single invariant representation. Here, the mean same-character similarity is $0.857$, the mean different-character similarity is $0.298$, and the corresponding gap is $0.559$.
This pattern indicates that the identity branch preserves static character identity, while the context branch retains additional sentence-dependent modulation.

\paragraph{Context Memory.}
Figure~\ref{fig:supp_probe_pca} further visualizes Context Memory using a 3D PCA projection.
Tokens with different character identities still occupy different regions, but repeated instances of the same character are dispersed rather than collapsed to one point.
Moreover, when colored by token order ($0 \rightarrow N\!-\!1$), the Context Memory tokens show a gradual movement trend along the sequence.
This indicates that the representation carries not only character identity but also context-dependent variation associated with token position and local word structure.
Taken together, these observations support the intended role split of the Character Context Encoder: the Character-Identity Embedding specifies \emph{what} to write, while Context Memory modulates \emph{how} it is realized and connected in sentence context.

\subsection{Bigram-Aware Local Decoding}
\label{sec:appendix_bigram_local_decoding}

\paragraph{Bi-SWT Decoder.}

The handwriting decoder uses a bigram-aware sliding-window design to model predecessor-conditioned local generation.
In the reported setting, the decoder operates with hidden dimension $512$ and $8$ attention heads, and it combines writer-style, glyph-style, and context-conditioned memories through separate decoder branches.
The sliding-window rule is \texttt{N\_GRAM\_AWARE\_SLIDING\_WINDOW=2}, which corresponds to a bigram prefix.
This design gives the decoder explicit access to the immediately preceding character when generating the current character trajectory.
As a result, the model can model local boundary phenomena such as cursive continuation, pen-lift transitions, and boundary-local spacing more reliably than a character-isolated decoder.
The predecessor-conditioned window is especially important under sparse sentence-level transition coverage, where many character pairs are not observed frequently enough to be learned robustly from sentence data alone.
During training, the bigram stream is introduced in Stage~2 and retained in Stage~3 together with the sentence stream.
This preserves local transition modeling even after full sentence supervision is activated.
In this sense, the Bi-SWT decoder and the curriculum design are tightly coupled: the decoder provides an explicit local transition mechanism, while the curriculum ensures that this mechanism is learned before sentence-level composition becomes dominant.

\paragraph{Gated Context Fusion.}

Context Memory improves inter-character connectivity, but excessive context injection can weaken writer/glyph stylization.
To balance these effects, \textsc{BoundInk} uses token-wise gated context fusion.

Let
\[
\mathbf{c}_t
=
\mathbf{m}^{\mathrm{ctx}}_{s(t)}
\]
denote the position-specific context vector for token $t$, and let
$\widetilde{\mathbf{h}}^{\mathrm{ctx}}_t$ denote the context-decoder output obtained by conditioning the style-conditioned decoder state
$\mathbf{h}^{\mathrm{sty}}_t$ on $\mathbf{c}_t$.
We compute the scalar gate as
\begin{equation}
g_t
=
\gamma\,
\sigma\!\left(
f_{\mathrm{gate}}\!\left(
\mathbf{h}^{\mathrm{sty}}_t
\right)
\right),
\label{eq:appendix_gate}
\end{equation}
and fuse the style-conditioned and context-conditioned states as
\begin{equation}
\mathbf{h}_t
=
(1-g_t)\mathbf{h}^{\mathrm{sty}}_t
+
g_t\widetilde{\mathbf{h}}^{\mathrm{ctx}}_t.
\label{eq:appendix_gate_fusion}
\end{equation}
Here $g_t\in(0,\gamma)$ is a scalar gate broadcast over the feature dimension,
$\gamma\in(0,1]$ is the gate cap, and $s(t)$ maps token $t$ to its character position.
The gate controls how much sentence context is injected at each token, enabling local connectivity adaptation without overwriting writer/glyph style cues.

\begin{figure}[t!]
    \centering
    \includegraphics[
        width=0.90\columnwidth
    ]{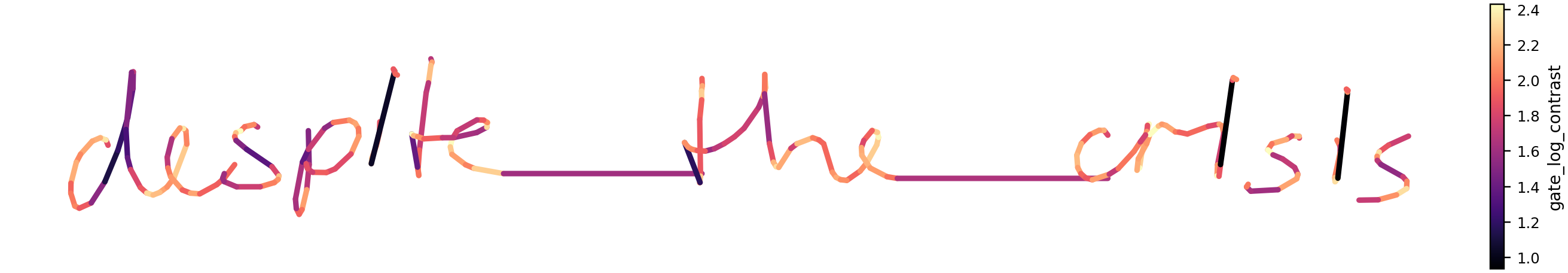}
    \caption{Token-wise responses of gated context fusion overlaid on a
    generated sentence trajectory.
    Higher responses occur near character boundaries and cursive connections,
    indicating stronger use of Context Memory in boundary-sensitive regions.}
    \label{fig:appendix_gated_context_response}
\end{figure}

The Bi-SWT decoder focuses on predecessor-conditioned local synthesis and handles boundary-local structure through bigram windows.
Longer-range sentence information is handled separately through Context Memory and gated context fusion.
The two modules therefore play complementary roles.
Bi-SWT captures local transition structure, while gated fusion injects context-dependent modulation on top of that local structure.
% Figure~\ref{fig:main_gate_overlay_logcontrast_clean} shows that the gate response is not uniform along the trajectory.
% Higher responses are concentrated near character boundaries and cursive-connection regions, whereas many ordinary intra-character stroke regions exhibit lower responses.
% This spatial pattern indicates that context contribution increases where local realization depends more strongly on neighboring characters.
% The same figure also shows that long spacing regions usually have weaker gate responses than boundary-local trajectory segments.
% In addition, characters whose local form changes little between isolated and sentence contexts often show relatively small gate values.
% For example, the character \textit{i} often changes less across contexts than highly connected cursive characters, and its gate responses are correspondingly weaker.
% These observations indicate that the gate varies with local context demand rather than remaining constant across the whole sentence.
% The sentence-level ablation in the main paper supports this interpretation.
% Removing gated context fusion preserves part of the DTW~\cite{berndt1994using} benefit but reduces CSM, with the largest drops appearing in $\mathrm{F1}_{\mathrm{cursive}}$ and KGS.
Figure~\ref{fig:appendix_gated_context_response} shows that the gate
response varies along the generated trajectory.
Higher responses are concentrated near character boundaries and cursive
connections, whereas many intra-character regions receive weaker context
contributions.
This pattern is consistent with the main-paper ablation, where removing
gated context fusion primarily reduces
$\mathrm{F1}_{\mathrm{Cursive}}$ and KGS.
This indicates that local bigram modeling alone is not sufficient and that sentence-conditioned context modulation provides an additional benefit for realistic boundary behavior and spacing.

\paragraph{GMM Output Head.}

Trajectory generation is modeled by a Gaussian mixture density head with $K=20$ components.
For each component, the decoder predicts the parameters
\begin{equation}
(\pi,\mu_x,\mu_y,\sigma_x,\sigma_y,\rho),
\end{equation}
where $\pi$ denotes the mixture weight, $(\mu_x,\mu_y)$ the mean, $(\sigma_x,\sigma_y)$ the standard deviations, and $\rho$ the correlation coefficient.
Following stable parameterization practice, we use
\begin{equation}
\sigma = \mathrm{softplus}(\cdot)+\epsilon,
\qquad
\rho = \tanh(\cdot),
\end{equation}
with $\rho$ additionally clamped to $\left[-(1-10^{-5}),\,1-10^{-5}\right]$.
This prevents numerical instability in the bivariate Gaussian covariance near $|\rho|=1$.
This output head is shared across curriculum stages and is optimized through the masked GMM negative log-likelihood term defined in Section~\ref{sec:appendix_training_details}.

% =========================================================
% Additional Experimental Analysis
% =========================================================
\section{Additional Experimental Analysis}
\label{sec:appendix_additional_analysis}

\subsection{Exploratory Sequence Style Reference and Vertical Stability}
\label{sec:appendix_vertical_drift_analysis}

\paragraph{Motivation and Evaluation Setting.}
The main \textsc{BoundInk} model uses image-based character references for
style conditioning.
This reference setting is practical for downstream use because it does not
require collecting writer-specific online trajectory sequences at inference
time.
Accordingly, the main model is designed to reproduce writer style and
boundary behavior from readily available appearance references while
explicitly modeling local predecessor--current transitions.

As an additional exploratory study, we investigate whether providing a
trajectory sequence as an extra style reference can further stabilize
long-horizon sentence generation.
Although character-image references provide strong local appearance cues,
they do not directly expose how a writer's geometry evolves across a
continuous sequence, including baseline progression, relative character
height, spacing, and boundary transitions.
These properties become particularly important during fully autoregressive
generation, where small local errors can accumulate across characters and
words and produce sentence-level artifacts such as vertical drift.

To study this setting, we construct an extended revision, denoted V2, that
augments the image-based references with a trajectory-based
\emph{Sequence Style Reference}.
This additional reference provides writer-specific sequence-level geometric
cues complementary to the appearance information used by the main model.
V2 further introduces an auxiliary boundary-state prediction objective that
encourages the decoder representation to retain explicit information about
spacing, connectivity, relative scale, and vertical boundary geometry.
Together with the existing Vertical Drift Loss (VDL), these additions target
sentence-generation stability through complementary conditioning,
representation regularization, and direct geometric supervision.

We evaluate this extension by focusing on \emph{free-running vertical drift},
a long-horizon artifact in which the vertical placement of generated words
becomes unstable across a sentence.
Unlike VDL, which provides adjacent-character supervision during training,
the free-running diagnostic is computed after complete autoregressive
generation and therefore captures accumulated sentence-level instability.

For this analysis, we compare two internal model revisions, V1 and V2, at
the same training budget of $500$k iterations.
Both are evaluated on the same fixed English test list under fully
autoregressive inference.
V1 retains the image-based reference setting, whereas V2 additionally uses
the trajectory-based Sequence Style Reference and boundary-aware auxiliary
regularization.
The $500$k checkpoint corresponds to approximately one half of the planned
$1$M-iteration training schedule and is used to examine intermediate
stabilization behavior rather than final converged performance.

Importantly, V2 also differs from V1 in several accompanying preprocessing
and training choices.
The comparison should therefore be interpreted as an exploratory
revision-level analysis rather than a controlled ablation that isolates the
effect of Sequence Style Reference or boundary-state regularization.

\paragraph{Sequence Style Reference.}
Let the trajectory reference sequence be
\begin{equation}
\mathcal{R}
=
\left\{
\left(
u_r,
\mathbf{Y}_r,
\mathbf{b}_r
\right)
\right\}_{r=1}^{R},
\end{equation}
where $u_r$ denotes the identity of reference character $r$,
$\mathbf{Y}_r=\{\mathbf{y}_{r,t}\}_{t=1}^{T_r}$ is its online trajectory,
and $\mathbf{b}_r$ denotes its boundary/layout features.
Each trajectory point uses the same six-dimensional representation as the
generator, consisting of two coordinates and four pen states
(\PM, \PU, \CEOC, and \EOC).

Each point is first mapped to the decoder feature dimension using a learned
projection $f_{\mathrm{pt}}$.
The variable-length trajectory is then summarized by masked mean pooling:
\begin{equation}
\bar{\mathbf{z}}_r
=
\frac{1}{T_r}
\sum_{t=1}^{T_r}
f_{\mathrm{pt}}
\left(
\mathbf{y}_{r,t}
\right).
\label{eq:appendix_seqref_traj}
\end{equation}

The trajectory summary is combined with a character-identity embedding and
a projected boundary representation:
\begin{equation}
\mathbf{q}_r
=
\operatorname{LN}
\left(
\bar{\mathbf{z}}_r
+
\mathbf{e}^{\mathrm{char}}(u_r)
+
f_{\mathrm{bnd}}(\mathbf{b}_r)
\right).
\label{eq:appendix_seqref_token}
\end{equation}
Boundary features are applied only at valid boundary positions.

A lightweight Transformer encoder contextualizes these reference tokens:
\begin{equation}
\mathbf{M}^{\mathrm{seq}}
=
\operatorname{TransEnc}_{\mathrm{seq}}
\left(
\mathbf{q}_{1:R}
\right),
\label{eq:appendix_seqref_memory}
\end{equation}
yielding a variable-length sequence-style memory
$\mathbf{M}^{\mathrm{seq}}\in\mathbb{R}^{R\times D}$.

The boundary representation contains nine geometric attributes:
horizontal advance, kerning gap, baseline shift, log height ratio,
log width ratio, cursive-join state, post-space horizontal displacement,
post-space vertical displacement, and the relative height of the first
character after a space.
Thus, the Sequence Style Reference represents not only individual trajectory
shapes but also writer-specific geometric relationships observed across the
reference sequence.

During generation, let $\mathbf{h}^{\mathrm{glyph}}_t$ denote the
glyph-conditioned decoder state at token $t$.
Cross-attention to the sequence-style memory produces
\begin{equation}
\widetilde{\mathbf{h}}^{\mathrm{seq}}_t
=
\operatorname{Dec}_{\mathrm{seq}}
\left(
\mathbf{h}^{\mathrm{glyph}}_t,
\mathbf{M}^{\mathrm{seq}}
\right).
\label{eq:appendix_seqref_decode}
\end{equation}
The resulting sequence-conditioned state is introduced through gated
residual interpolation:
\begin{equation}
\mathbf{h}^{\mathrm{seq}}_t
=
\mathbf{h}^{\mathrm{glyph}}_t
+
g_{\mathrm{seq}}
\left(
\widetilde{\mathbf{h}}^{\mathrm{seq}}_t
-
\mathbf{h}^{\mathrm{glyph}}_t
\right),
\label{eq:appendix_seqref_fusion}
\end{equation}
where
\begin{equation}
g_{\mathrm{seq}}
=
\gamma_{\mathrm{seq}}
\sigma(a_{\mathrm{seq}}).
\label{eq:appendix_seqref_gate}
\end{equation}
For the evaluated V2 configuration,
$\gamma_{\mathrm{seq}}=0.35$.
This cap limits the influence of the trajectory reference so that
sequence-level cues supplement rather than overwrite the existing
writer/glyph representation.
The resulting state is subsequently processed by the sentence-context
pathway of \textsc{BoundInk}.

The same boundary representation used to construct Sequence Style Reference
tokens also serves as an auxiliary supervision target.
Thus, V2 uses boundary geometry both as reference-side conditioning and as
decoder-side regularization.

\paragraph{Boundary-State Auxiliary Objective.}
In addition to sequence-level conditioning, V2 uses an auxiliary multi-task
objective to encourage the shared decoder representation to retain explicit
boundary and layout information.

For character position $s$, the target boundary state is
\begin{equation}
\mathbf{b}_s
=
\left[
a_s,\,
k_s,\,
\delta^{\mathrm{base}}_s,\,
r^h_s,\,
r^w_s,\,
j_s,\,
d^{x,\mathrm{space}}_s,\,
d^{y,\mathrm{space}}_s,\,
r^{h,\mathrm{post}}_s
\right],
\label{eq:appendix_boundary_state}
\end{equation}
where $a_s$ is horizontal advance,
$k_s$ is the kerning gap,
$\delta^{\mathrm{base}}_s$ is the baseline shift,
$r^h_s$ and $r^w_s$ are log height and width ratios,
$j_s$ is the cursive-join indicator,
$d^{x,\mathrm{space}}_s$ and $d^{y,\mathrm{space}}_s$ represent
post-space horizontal and vertical displacement, and
$r^{h,\mathrm{post}}_s$ represents the relative height of the first
character following a space.

A lightweight prediction head estimates this state from the decoder
representation associated with each character:
\begin{equation}
\widehat{\mathbf{b}}_s
=
f_{\mathrm{aux}}
\left(
\mathbf{h}_s
\right),
\label{eq:appendix_boundary_pred}
\end{equation}
where $f_{\mathrm{aux}}$ consists of layer normalization followed by a
two-layer MLP.

The eight continuous attributes are optimized with mean squared error:
\begin{equation}
\mathcal{L}_{\mathrm{bnd}}^{\mathrm{cont}}
=
\frac{1}{|\mathcal{V}|}
\sum_{s\in\mathcal{V}}
\left\|
\widehat{\mathbf{b}}^{\,\mathrm{cont}}_s
-
\mathbf{b}^{\mathrm{cont}}_s
\right\|_2^2,
\label{eq:appendix_boundary_cont}
\end{equation}
where $\mathcal{V}$ denotes valid boundary positions.

For the binary cursive-join component, we use
\begin{equation}
\mathcal{L}_{\mathrm{bnd}}^{\mathrm{join}}
=
\frac{1}{|\mathcal{V}|}
\sum_{s\in\mathcal{V}}
\operatorname{BCEWithLogits}
\left(
\hat{j}_s,
j_s
\right).
\label{eq:appendix_boundary_join}
\end{equation}
The complete boundary-state auxiliary objective is
\begin{equation}
\mathcal{L}_{\mathrm{bnd}}
=
\mathcal{L}_{\mathrm{bnd}}^{\mathrm{cont}}
+
\mathcal{L}_{\mathrm{bnd}}^{\mathrm{join}}.
\label{eq:appendix_boundary_loss}
\end{equation}

In the evaluated V2 configuration, the auxiliary term is weighted by
\begin{equation}
\lambda_{\mathrm{bnd}} = 0.003.
\end{equation}
The predicted boundary state is not fed back into autoregressive trajectory
generation at inference time.
Instead, it acts as a training-time regularizer that encourages the shared
decoder representation to preserve explicit information about baseline
displacement, relative character scale, spacing, and connectivity.

\paragraph{Complementary Geometric Supervision.}
Sequence Style Reference, boundary-state auxiliary regularization, and VDL
address sentence geometry at different levels.

Sequence Style Reference provides observed writer-specific geometric cues
from a trajectory sequence.
Boundary-state prediction encourages the character-level decoder
representation to explicitly encode local layout properties.
VDL directly penalizes disagreement in adjacent-character vertical geometry
during training.

Specifically, VDL compares the predicted and ground-truth changes in top,
centroid, and bottom statistics between consecutive characters.
It therefore provides local geometric supervision, whereas the trajectory
reference provides longer-range writer-specific cues and the auxiliary task
regularizes the latent representation.
These three mechanisms are complementary rather than redundant.

\paragraph{Free-Running Vertical Drift Diagnostic.}
We next examine whether the V2 revision exhibits improved vertical stability
under fully autoregressive sentence generation.
Unlike VDL, this diagnostic is computed after complete decoding, where local
trajectory errors can accumulate across characters and words.

For sentence $i$, let
\begin{equation}
\mathcal{W}_i
=
\left(
W_{i,1},
\ldots,
W_{i,M_i}
\right)
\end{equation}
denote its ordered sequence of words.
The vertical center of word $m$ is defined as
\begin{equation}
c_{i,m}
=
\operatorname{median}
\left\{
y_{s,t}
\;\middle|\;
u_s \in W_{i,m}
\right\},
\label{eq:appendix_word_center}
\end{equation}
using trajectory points belonging to the non-space characters in that word.

For each non-space character $s$, its vertical extent is
\begin{equation}
h_{i,s}
=
\max_t y_{s,t}
-
\min_t y_{s,t},
\label{eq:appendix_char_height}
\end{equation}
and the sentence-level vertical scale is
\begin{equation}
H_i
=
\operatorname{median}_{s:\,h_{i,s}>0}
h_{i,s}.
\label{eq:appendix_sentence_height}
\end{equation}

We define the free-running vertical drift ratio as
\begin{equation}
R_i^{\mathrm{drift}}
=
\frac{
\max_{1\leq m<M_i}
\left|
c_{i,m+1}
-
c_{i,m}
\right|
}{
H_i
}.
\label{eq:appendix_free_running_drift}
\end{equation}
The numerator measures the largest vertical displacement between adjacent
word centers, while normalization by $H_i$ reduces sensitivity to
writer-specific character scale.
Lower values indicate greater vertical stability.

To measure agreement with the corresponding real handwriting, we also
report
\begin{equation}
E_{\mathrm{drift}}
=
\frac{1}{N}
\sum_{i=1}^{N}
\left|
\widehat{R}_i^{\mathrm{drift}}
-
R_i^{\mathrm{drift}}
\right|,
\label{eq:appendix_drift_mae}
\end{equation}
where $\widehat{R}_i^{\mathrm{drift}}$ is measured from the generated
sentence and $R_i^{\mathrm{drift}}$ from its ground truth.

To avoid assigning predicted trajectory segments to incorrect word
positions, we retain only sentences for which prediction and ground truth
admit an exact one-to-one alignment between end-of-character segments and
target characters.
No heuristic segment splitting or padding is applied.
This yields a common strictly aligned subset of $276$ sentences from
$57$ writers.

The threshold
$R_i^{\mathrm{drift}}>1.20$
is retained from the diagnostic criterion used during fixed-test-list
curation.
Under this criterion, none of the retained ground-truth sentences exceeds
the threshold.

\begin{table}[t!]
\centering
\small
\setlength{\tabcolsep}{3.0pt}
\renewcommand{\arraystretch}{1.0}

\begin{tabular}{@{}lrrrr@{}}
\toprule
\textbf{Model} &
\textbf{Mean $\downarrow$} &
\textbf{Median $\downarrow$} &
\textbf{$>1.20$ $\downarrow$} &
\textbf{GT MAE $\downarrow$} \\
\midrule
Ground truth
    & 0.3015
    & 0.2572
    & 0.0\%
    & -- \\
V1 ($500$k)
    & 0.6999
    & 0.5408
    & 14.5\%
    & 0.4853 \\
V2 ($500$k)
    & \textbf{0.4994}
    & \textbf{0.4179}
    & \textbf{4.0\%}
    & \textbf{0.3065} \\
\bottomrule
\end{tabular}

\caption{Intermediate-checkpoint free-running vertical drift on the common
strictly aligned subset of $276$ sentences from $57$ writers.
V2 incorporates trajectory-based Sequence Style Reference conditioning and
boundary-state auxiliary regularization together with other revision-level
changes.
Both revisions are evaluated at $500$k iterations.
Mean and median are computed from
Eq.~\eqref{eq:appendix_free_running_drift}, and GT MAE follows
Eq.~\eqref{eq:appendix_drift_mae}.
Lower values are better.}
\label{tab:appendix_vertical_drift_checkpoints}
\end{table}

\begin{figure*}[t!]
    \centering
    \includegraphics[
        width=0.9\textwidth
    ]{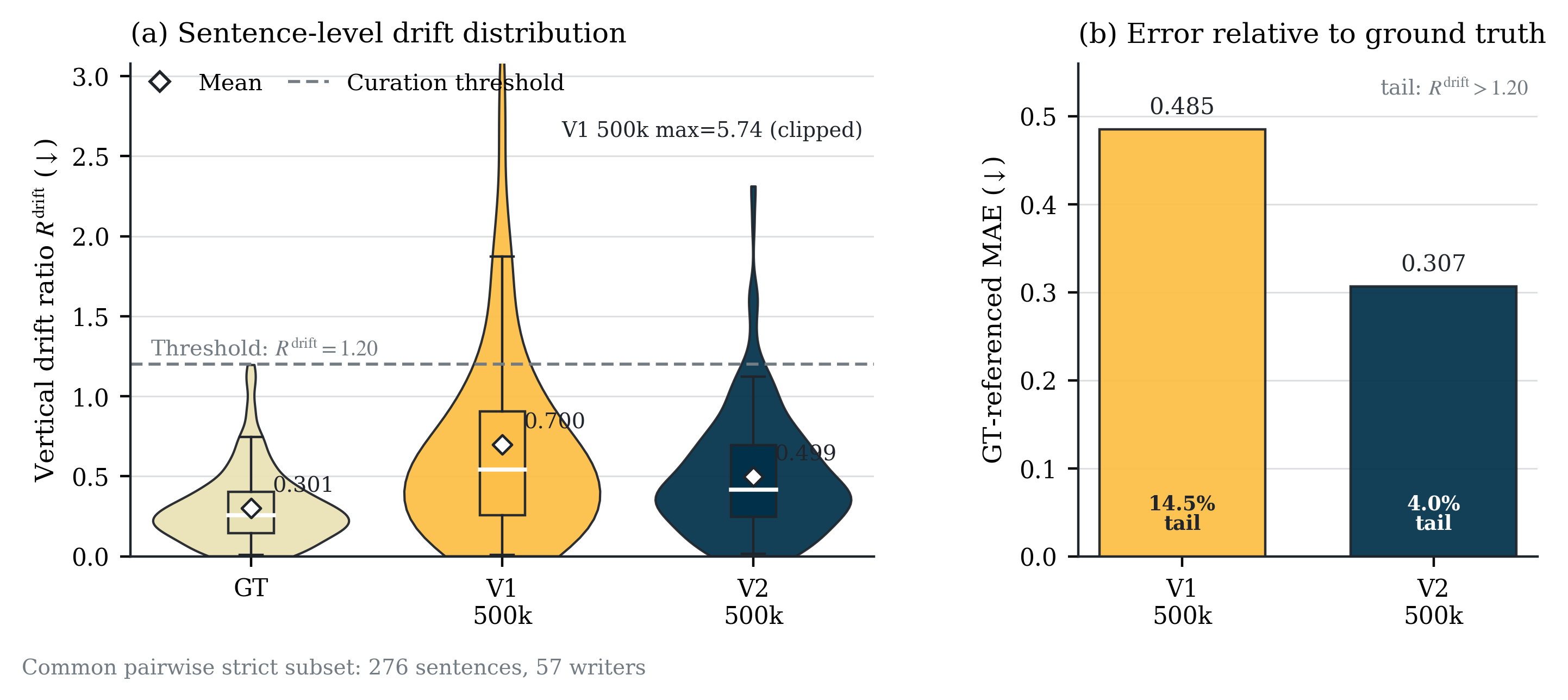}

    \caption{Free-running vertical drift at the matched $500$k checkpoint.
    Panel~(a) shows sentence-level drift-ratio distributions.
    Internal boxes span the interquartile range, white lines indicate medians,
    diamonds indicate means, whiskers span the 5th--95th percentiles, and the
    dashed line marks the $1.20$ diagnostic threshold.
    Panel~(b) reports mean absolute error relative to the ground-truth drift
    ratio; labels inside the bars indicate the percentage of sentences above
    $1.20$.
    Lower values are better in both panels.
    V2 includes sequence-level trajectory conditioning and boundary-aware
    auxiliary regularization, but the comparison is revision-level rather
    than a single-component ablation.}
    \label{fig:appendix_vertical_drift_checkpoints}
\end{figure*}

Figure~\ref{fig:appendix_vertical_drift_checkpoints} visualizes the
sentence-level distributions corresponding to
Table~\ref{tab:appendix_vertical_drift_checkpoints}.
The distribution view complements the aggregate statistics by exposing both
typical vertical drift and the high-drift tail.

At the matched training horizon, the mean free-running drift ratio decreases
from $0.6999$ for V1 to $0.4994$ for V2, corresponding to a $28.7\%$
reduction.
The ground-truth-referenced drift error decreases from $0.4853$ to $0.3065$,
a $36.8\%$ reduction, while the proportion of severe-drift sentences above
$1.20$ decreases from $14.5\%$ to $4.0\%$.
The writer-level bootstrap $95\%$ confidence intervals for the
V2-minus-V1 differences are $[-0.3234,-0.0676]$ for the raw drift ratio
and $[-0.2713,-0.0582]$ for the ground-truth-referenced error.
Both intervals exclude zero.

These results indicate that V2 develops substantially greater vertical
stability on the strictly aligned subset at the matched intermediate
checkpoint.
The observed behavior is consistent with the motivation behind the V2
extension: trajectory-level references provide information about
writer-specific baseline progression, relative height, spacing, and boundary
geometry, while auxiliary boundary prediction encourages the decoder
representation to preserve these properties.

However, this experiment does not isolate the causal contribution of either
Sequence Style Reference or boundary-state auxiliary regularization.
The improvement should therefore be interpreted as a property of the complete
V2 revision rather than as a single-module ablation.

\paragraph{Results and Limitations.}
This experiment is intended as a targeted diagnostic of sentence-level
vertical stability and does not replace CSM or DTW.

First, V1 and V2 are evaluated at an intermediate $500$k checkpoint rather
than at the end of the planned $1$M-iteration training schedule.
Their mean generated drift ratios, $0.6999$ for V1 and $0.4994$ for V2,
remain above the ground-truth value of $0.3015$.

Second, the strictly aligned subset contains only $276$ of the $588$
generated test sentences because exact character segmentation is required
for reliable assignment of trajectory segments to words.
The analysis is therefore conditional on successful segmentation and may
under-represent severe generation failures.

Third, V2 is a model revision rather than a controlled single-factor
ablation.
In addition to Sequence Style Reference and the boundary-state auxiliary
objective, V2 differs from V1 in preprocessing, space handling, boundary
representation, loss configuration, and Stage~3 training.
Accordingly, the current comparison cannot determine how much of the
observed vertical stabilization is attributable to each individual change.

Finally, on all $588$ paired sentences, normalized DTW at $500$k is lower
for V1 ($1.3152$) than for V2 ($2.6205$).
These values are also intermediate-checkpoint diagnostics and should not be
interpreted as final model performance.
The result emphasizes that improved vertical stability represents one
specific geometric property and does not imply uniformly better trajectory
similarity at this stage of training.

Taken together, this analysis suggests that sequence-level trajectory
conditioning and boundary-aware auxiliary supervision are promising
complements to the local geometric supervision provided by VDL.
Controlled ablations at converged checkpoints are required to isolate the
individual contribution of each V2 component.

% =========================================================
% Human Evaluation Protocol
% =========================================================
\section{Human Evaluation Protocol}
\label{sec:appendix_human_eval}

\subsection{Human Evaluation Design and Interface}
\label{sec:appendix_human_study}

\paragraph{Experimental Design.}

To complement the quantitative evaluation, we conducted a blind A/B study
comparing \textsc{BoundInk} with
\textsc{DeepWriting}~\cite{aksan2018deepwriting} and
\textsc{DSD}~\cite{kotani2020generating} in terms of writer-style fidelity,
inter-character connectivity, and spacing.

We prepared five questionnaire versions (V1--V5), each containing 20 comparison items.
Protocol assignment was balanced within each version: 10 items compared \textsc{BoundInk} against \textsc{DSD}~\cite{kotani2020generating}, and the other 10 compared \textsc{BoundInk} against \textsc{DeepWriting}~\cite{aksan2018deepwriting}.
Each item was judged separately for Style, Connectivity, and Spacing, resulting in 60 criterion-level decisions per completed submission.
Each item consisted of three real reference images written by the same
writer and two generated candidates for the same target sentence.
The two candidates corresponded to \textsc{BoundInk} and one baseline and
were shown as Candidate A and Candidate B with randomized left/right
assignment.
The reference images were sampled from three different sentences written
by the same writer.
All online trajectories were rendered as PNG images and scaled to a common
height for presentation.
For fair visual comparison, the three reference images and the two generated candidates are scaled to the same height.
This normalization keeps the comparison focused on writer style, boundary connectivity, and spacing patterns rather than trivial differences in absolute image scale.

In total, 30 participants completed the study, with one submission per participant and six submissions for each questionnaire version.
Since each participant evaluated 20 items under three criteria, the study produced 1,800 criterion-level decisions.
Among them, 1,604 were valid preference decisions and 196 were marked as \emph{Cannot judge}, corresponding to an abstention rate of 10.9\%.
Multiple questionnaire versions serve two purposes.
They distribute the item pool across raters while keeping the per-rater burden moderate, and they maintain protocol balance between \textsc{BoundInk} vs.~\textsc{DSD}~\cite{kotani2020generating} and \textsc{BoundInk} vs.~\textsc{DeepWriting}~\cite{aksan2018deepwriting}.
Participants were Korean adults and included a mixed group of graduate students and working professionals.
Gender was not controlled and was naturally mixed across participants.
Participants were asked to choose which candidate better matched the reference writer under three criteria.
\textbf{Style} evaluates overall handwriting appearance, including slant, visual tone, rounded versus angular shape, and writing rhythm.
\textbf{Connectivity} evaluates inter-character connection patterns, including natural continuation or separation between characters and smooth boundary transitions.
\textbf{Spacing} evaluates spacing between words, including inter-word distance, spacing consistency, and whether gaps appear excessively narrow or wide.

% \begin{table*}[t]
% \centering
% \small
% \setlength{\tabcolsep}{6pt}
% \renewcommand{\arraystretch}{1.10}
% \begin{tabular}{@{} l >{\raggedright\arraybackslash}p{11.5cm} @{}}
% \toprule
% \textbf{Item} & \textbf{Value} \\
% \midrule
% Questionnaire versions & 5 balanced versions (V1--V5) \\
% Submissions & 30 total (6 submissions per version) \\
% Items per version & 20 randomized A/B comparison items \\
% Criteria per item & Style / Connectivity / Spacing \\
% Protocol balance & 10 DSD + 10 DeepWriting per version \\
% Total decisions & 1,800 criterion-level decisions \\
% Valid / Cannot judge & 1,604 valid decisions / 196 Cannot judge (10.9\% abstention rate) \\
% Sentence IDs used & 40 \\
% Unique sentence texts & 23 \\
% Sentence length over unique texts & min / median / mean / max = 12 / 16 / 16.18 / 19 \\
% \bottomrule
% \end{tabular}
% \caption{Summary of the human evaluation setup and collected responses.}
% \label{tab:appendix_human_eval_summary}
% \end{table*}

\paragraph{Questionnaire Interface.}

\begin{figure}[t]
    \centering
    \includegraphics[width=\columnwidth]{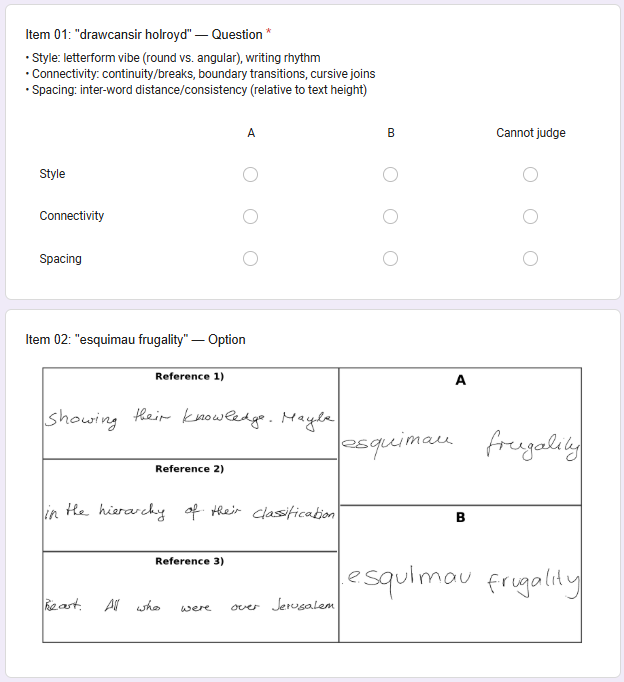}
    \caption{Actual questionnaire interface used in the human evaluation.
    Each item presents three reference samples and two randomized candidates
    evaluated under Style, Connectivity, and Spacing, with an additional
    \emph{Cannot judge} option.}
    \label{fig:appendix_human_eval_interface_real}
\end{figure}

Before the evaluation, participants received a brief explanation of the
Style, Connectivity, and Spacing criteria.
The actual evaluation items contained no explanatory annotations.

Figure~\ref{fig:appendix_human_eval_interface_real} shows the questionnaire
interface used to collect responses.
Each item presents three real handwriting samples from the reference writer
and two generated candidates for the same target sentence.
Candidate order is randomized, and participants record separate judgments
for Style, Connectivity, and Spacing, with an additional
\emph{Cannot judge} option.
This design helps reduce instructional bias during the actual response stage while preserving criterion-level evaluation.
The annotations are therefore shown only on the instruction page and are not included in the actual evaluation items.
This separation ensures that participants receive guidance on the intended criteria before the study begins, while the final judgments themselves are collected from an unannotated interface.

% =========================================================
% Connectivity and Spacing Metrics
% =========================================================
\section{Connectivity and Spacing Metrics (CSM)}
\label{sec:appendix_metrics}

This section provides the complete definitions of the four components of
Connectivity and Spacing Metrics (CSM):
$\mathrm{F1}_{\mathrm{Cursive}}$, CRE, KGS, and SSS.
The first two components evaluate cursive connectivity, whereas KGS and SSS
evaluate inter-character and inter-word spacing, respectively.
All reported values are aggregated at the writer level and then
macro-averaged across writers.

% ---------------------------------------------------------
% F.1 Boundary and Cursive Metrics
% ---------------------------------------------------------
\subsection{Boundary and Cursive Metrics}
\label{sec:appendix_boundary_cursive_metrics}
% Legacy label retained to avoid breaking existing references.
\label{sec:appendix_boundary_connectivity_metrics}

\paragraph{Boundary Formalization.}

\textsc{BoundInk} predicts four pen states at each trajectory step:
\[
\{\PM,\PU,\CEOC,\EOC\}.
\]
Here, \CEOC\ denotes an end-of-character event with the pen still down,
representing an explicit cursive continuation to the next character.
By contrast, \EOC\ denotes an end-of-character event without pen-down
continuation.

Some baseline methods do not predict an explicit \CEOC\ label.
For these methods, we first assign \EOC\ to each character boundary and
then infer cursive continuation using a distance-based heuristic.
Let $\mathbf{p}^{\mathrm{end}}_{s}$ be the last point of character $s$ and
let $\mathbf{p}^{\mathrm{PM}}_{s+1}$ be the first pen-move point of the
following character.
We define the connection distance as
\begin{equation}
d^{\mathrm{conn}}_s
=
\left\|
\mathbf{p}^{\mathrm{end}}_{s}
-
\mathbf{p}^{\mathrm{PM}}_{s+1}
\right\|_2 .
\label{eq:appendix_connection_distance}
\end{equation}
On normalized coordinates, we relabel the boundary as \CEOC\ when
\begin{equation}
d^{\mathrm{conn}}_s < \tau_{\mathrm{conn}},
\qquad
\tau_{\mathrm{conn}} = 0.005,
\label{eq:appendix_connection_threshold}
\end{equation}
and retain \EOC\ otherwise.
This heuristic is used only for baselines without explicit cursive-boundary
labels.
For \textsc{BoundInk}, the predicted boundary state is used directly.

Let $\mathcal{W}$ denote the set of writers and let $\mathcal{D}_w$ be the
set of test sentences for writer $w\in\mathcal{W}$.
A sentence $i\in\mathcal{D}_w$ consists of a character sequence
\[
\mathbf{x}^{(i)}
=
\left(
x^{(i)}_1,\ldots,x^{(i)}_{S_i}
\right),
\]
including spaces, and per-character trajectories
\[
\mathbf{Y}^{(i)}_s
=
\left\{
(x_{s,t},y_{s,t})
\right\}_{t=1}^{T^{(i)}_s}.
\]
We use the last trajectory step of character $s$ to define its ground-truth
boundary state
$q^{(i)}_s\in\{\PM,\PU,\CEOC,\EOC\}$ and its predicted counterpart
$\hat{q}^{(i)}_s$.
We use $\epsilon>0$, set to $10^{-6}$, for numerical stability.

Connectivity and local kerning are evaluated on adjacent non-space
character boundaries:
\begin{equation}
\mathcal{B}_i
=
\left\{
s
\;\middle|\;
\begin{aligned}
&1\le s<S_i,\\
&x^{(i)}_s\neq\texttt{space},\\
&x^{(i)}_{s+1}\neq\texttt{space}
\end{aligned}
\right\}.
\label{eq:appendix_adjacent_boundaries}
\end{equation}

Word spacing is evaluated on space-run boundaries:
\begin{equation}
\mathcal{P}_i
=
\left\{
(u,v)
\;\middle|\;
\begin{aligned}
&1\le u<v\le S_i,\\
&x^{(i)}_u\neq\texttt{space},\\
&x^{(i)}_v\neq\texttt{space},\\
&x^{(i)}_{u+1}
=\cdots=
x^{(i)}_{v-1}
=\texttt{space}
\end{aligned}
\right\}.
\label{eq:appendix_space_boundaries}
\end{equation}
Thus, a run of one or more spaces between two non-space characters is
counted as one word-spacing boundary.

For each character, we define its horizontal bounds as
\begin{gather}
L^{(i)}_s
=
\min_t x^{(i)}_{s,t},
\\
R^{(i)}_s
=
\max_t x^{(i)}_{s,t},
\end{gather}
and analogously define
$\hat{L}^{(i)}_s$ and $\hat{R}^{(i)}_s$ from the predicted trajectory.

\paragraph{$\mathrm{F1}_{\mathrm{Cursive}}$.}

We treat \CEOC\ as the positive class on each eligible boundary
$s\in\mathcal{B}_i$.
For \textsc{BoundInk}, the predicted state is read directly from the model
output.
For baselines without an explicit \CEOC\ state, the converted state from
Eqs.~\eqref{eq:appendix_connection_distance}
and~\eqref{eq:appendix_connection_threshold} is used.

We define the ground-truth and predicted binary connection labels as
\begin{gather}
z^{(i)}_s
=
\mathbb{I}
\left[
q^{(i)}_s=\CEOC
\right],
\\
\hat{z}^{(i)}_s
=
\mathbb{I}
\left[
\hat{q}^{(i)}_s=\CEOC
\right].
\end{gather}

Because F1 is non-linear, we aggregate boundary counts within each writer
rather than averaging sentence-level F1 values:
\begin{align}
\mathrm{TP}_w
&=
\sum_{i\in\mathcal{D}_w}
\sum_{s\in\mathcal{B}_i}
\mathbb{I}
\left[
\hat{z}^{(i)}_s=1
\wedge
z^{(i)}_s=1
\right],
\\
\mathrm{FP}_w
&=
\sum_{i\in\mathcal{D}_w}
\sum_{s\in\mathcal{B}_i}
\mathbb{I}
\left[
\hat{z}^{(i)}_s=1
\wedge
z^{(i)}_s=0
\right],
\\
\mathrm{FN}_w
&=
\sum_{i\in\mathcal{D}_w}
\sum_{s\in\mathcal{B}_i}
\mathbb{I}
\left[
\hat{z}^{(i)}_s=0
\wedge
z^{(i)}_s=1
\right].
\end{align}

Writer-level precision and recall are
\begin{gather}
\mathrm{Prec}_w
=
\frac{
\mathrm{TP}_w
}{
\mathrm{TP}_w+\mathrm{FP}_w+\epsilon
},
\\
\mathrm{Rec}_w
=
\frac{
\mathrm{TP}_w
}{
\mathrm{TP}_w+\mathrm{FN}_w+\epsilon
},
\end{gather}
and the writer-level cursive-boundary F1 score is
\begin{equation}
\mathrm{F1}^{\mathrm{Cursive}}_w
=
\frac{
2\,\mathrm{Prec}_w\,\mathrm{Rec}_w
}{
\mathrm{Prec}_w+\mathrm{Rec}_w+\epsilon
}.
\label{eq:f1_cursive}
\end{equation}

If neither the ground truth nor the prediction contains a positive cursive
boundary for writer $w$, we set
$\mathrm{F1}^{\mathrm{Cursive}}_w=1$ by convention.
At the protocol level, however, cursive-related metrics are reported as
undefined when the entire evaluation set contains no cursive-positive
boundaries, as in the OLHWG-compatible comparison.

\begin{figure}[t]
    \centering
    \includegraphics[
        width=\columnwidth
    ]{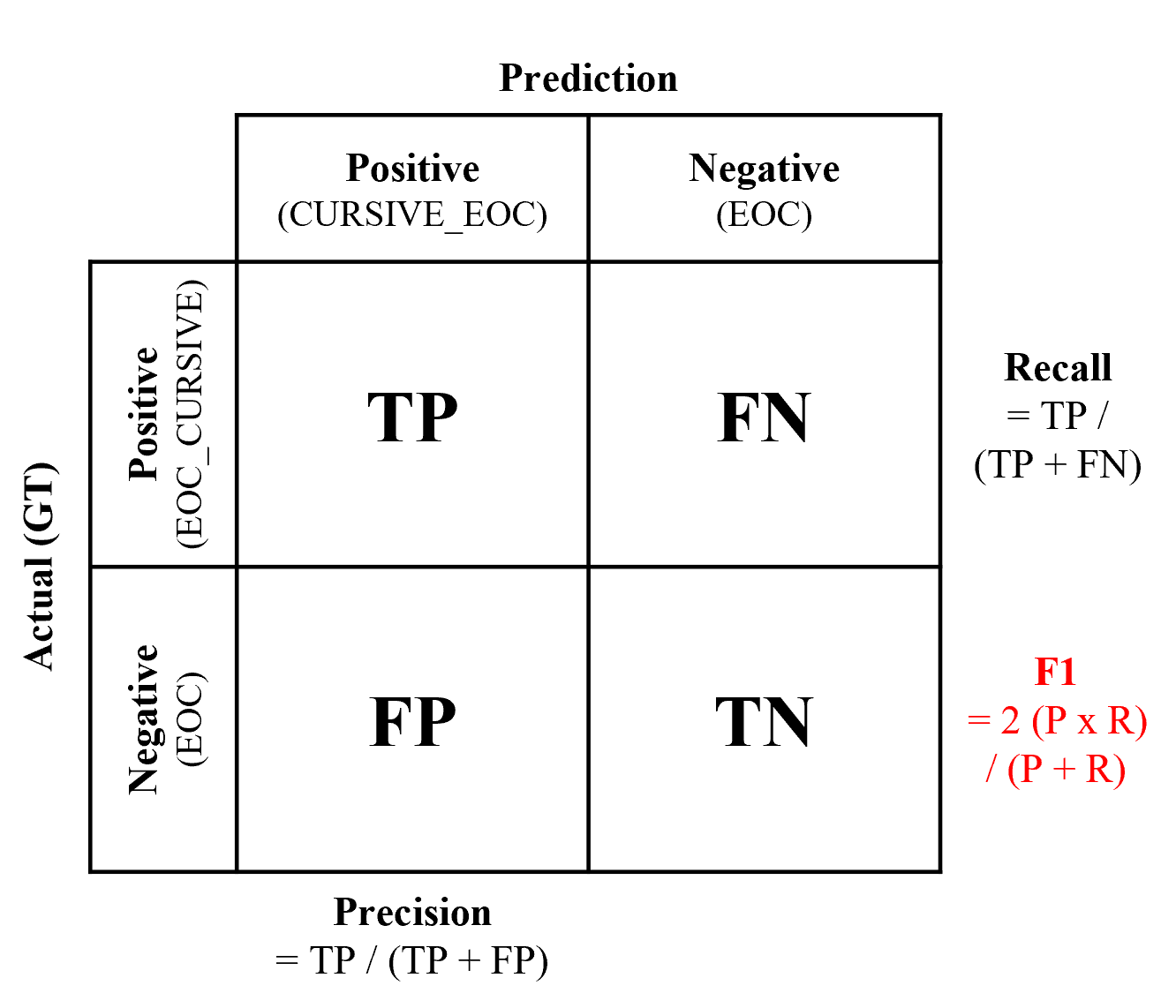}
    \caption{Illustrative example of
    $\mathrm{F1}_{\mathrm{Cursive}}$.
    Each eligible adjacent non-space boundary is classified as
    cursive-connected (\CEOC) or non-cursive (\EOC).}
    \label{fig:appendix_f1_cursive}
\end{figure}

Figure~\ref{fig:appendix_f1_cursive} illustrates the boundary-level
classification used by $\mathrm{F1}_{\mathrm{Cursive}}$.

\paragraph{CRE.}

$\mathrm{F1}_{\mathrm{Cursive}}$ evaluates individual boundary decisions,
but writers also differ in how frequently they use cursive continuation.
CRE therefore compares sentence-level cursive-continuation rates while
avoiding disproportionate weighting of longer sentences.

For sentence $i$, the ground-truth and predicted cursive rates are
\begin{gather}
r_i
=
\frac{
\sum_{s\in\mathcal{B}_i}
z^{(i)}_s
}{
|\mathcal{B}_i|+\epsilon
},
\\
\hat{r}_i
=
\frac{
\sum_{s\in\mathcal{B}_i}
\hat{z}^{(i)}_s
}{
|\mathcal{B}_i|+\epsilon
}.
\end{gather}
Sentences with $|\mathcal{B}_i|=0$ are retained and yield a rate of zero.

The writer-level mean absolute cursive-rate error is
\begin{equation}
\mathrm{MAE}^{\mathrm{rate}}_w
=
\frac{1}{|\mathcal{D}_w|}
\sum_{i\in\mathcal{D}_w}
\left|
\hat{r}_i-r_i
\right|,
\end{equation}
and CRE is defined as
\begin{equation}
\mathrm{CRE}_w
=
\max
\left(
0,\,
1-\mathrm{MAE}^{\mathrm{rate}}_w
\right).
\label{eq:cre}
\end{equation}

\begin{figure}[t]
    \centering
    \includegraphics[
        width=\columnwidth
    ]{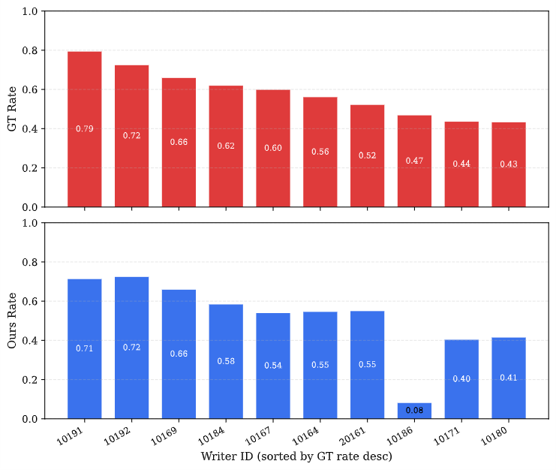}
    \caption{Illustrative example of CRE.
    CRE compares sentence-level cursive-continuation rates and aggregates
    their absolute differences at the writer level.}
    \label{fig:appendix_cre}
\end{figure}

Figure~\ref{fig:appendix_cre} illustrates the writer-level rate comparison.
In the shown example, the mean absolute rate difference is $0.0656$, which
gives $\mathrm{CRE}=0.9344$.

% ---------------------------------------------------------
% F.2 Spacing Metrics
% ---------------------------------------------------------
\subsection{Spacing Metrics (KGS and SSS)}
\label{sec:appendix_spacing_metrics}

\paragraph{KGS.}

For each adjacent non-space boundary $s\in\mathcal{B}_i$, we define the
ground-truth and predicted kerning gaps as
\begin{gather}
g^{(i)}_s
=
L^{(i)}_{s+1}
-
R^{(i)}_s,
\\
\hat{g}^{(i)}_s
=
\hat{L}^{(i)}_{s+1}
-
\hat{R}^{(i)}_s.
\end{gather}

A negative gap indicates overlap between adjacent characters.
We therefore use the non-negative component
$g^{+}=\max(g,0)$ together with an overlap penalty.
Let $\rho\in(0,1]$ and set $\rho=0.5$:
\begin{equation}
\pi(g,\hat{g})
=
\rho^{
\mathbb{I}
\left[
g<0
\;\vee\;
\hat{g}<0
\right]
}.
\label{eq:appendix_overlap_penalty}
\end{equation}

KGS uses an overlap-aware symmetric log-ratio similarity:
\begin{equation}
\begin{split}
\mathrm{KGS}^{(i)}_s
&=
\pi
\left(
g^{(i)}_s,
\hat{g}^{(i)}_s
\right)
\\
&\quad\cdot
\exp
\left(
-
\left|
\log
\frac{
\max(\hat{g}^{(i)}_s,0)+\epsilon
}{
\max(g^{(i)}_s,0)+\epsilon
}
\right|
\right).
\end{split}
\label{eq:kgs_pair}
\end{equation}

\begin{figure}[t]
    \centering
    \includegraphics[
        width=\columnwidth
    ]{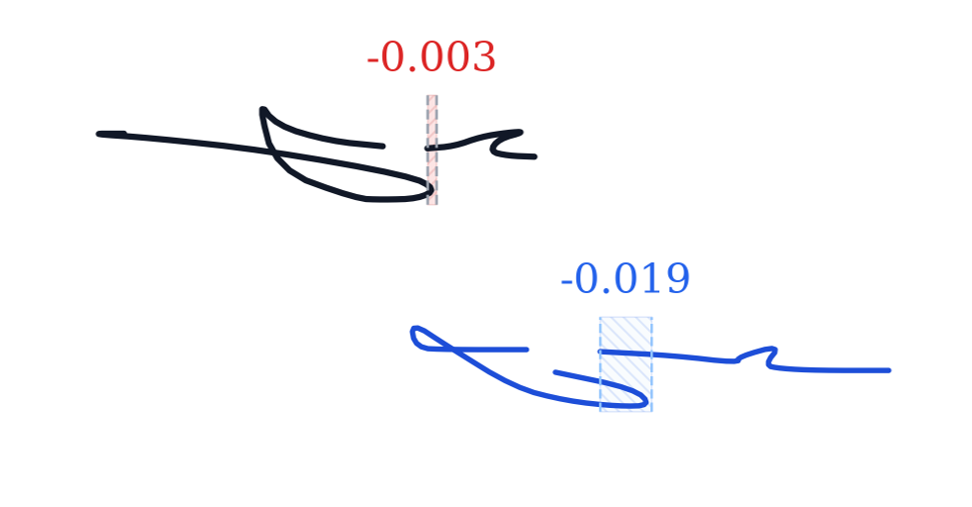}
    \caption{Illustrative example of KGS.
    KGS compares adjacent-character horizontal gaps using an overlap-aware
    symmetric log-ratio similarity.}
    \label{fig:appendix_kgs}
\end{figure}

Figure~\ref{fig:appendix_kgs} illustrates this comparison.
When either the ground-truth or predicted pair overlaps, the overlap penalty
is activated.
In the illustrated case, both trajectories overlap and the resulting KGS
value is $0.5$.

\paragraph{SSS.}

For each word-spacing boundary $(u,v)\in\mathcal{P}_i$, we define the
ground-truth and predicted spacing widths as
\begin{gather}
d^{(i)}_{u\rightarrow v}
=
L^{(i)}_v
-
R^{(i)}_u,
\\
\hat{d}^{(i)}_{u\rightarrow v}
=
\hat{L}^{(i)}_v
-
\hat{R}^{(i)}_u.
\end{gather}

SSS uses the same overlap-aware symmetric log-ratio form:
\begin{equation}
\begin{split}
\mathrm{SSS}^{(i)}_{u\rightarrow v}
&=
\pi
\left(
d^{(i)}_{u\rightarrow v},
\hat{d}^{(i)}_{u\rightarrow v}
\right)
\\
&\quad\cdot
\exp
\left(
-
\left|
\log
\frac{
\max(\hat{d}^{(i)}_{u\rightarrow v},0)+\epsilon
}{
\max(d^{(i)}_{u\rightarrow v},0)+\epsilon
}
\right|
\right).
\end{split}
\label{eq:sss_pair}
\end{equation}

\begin{figure}[t]
    \centering
    \includegraphics[
        width=\columnwidth
    ]{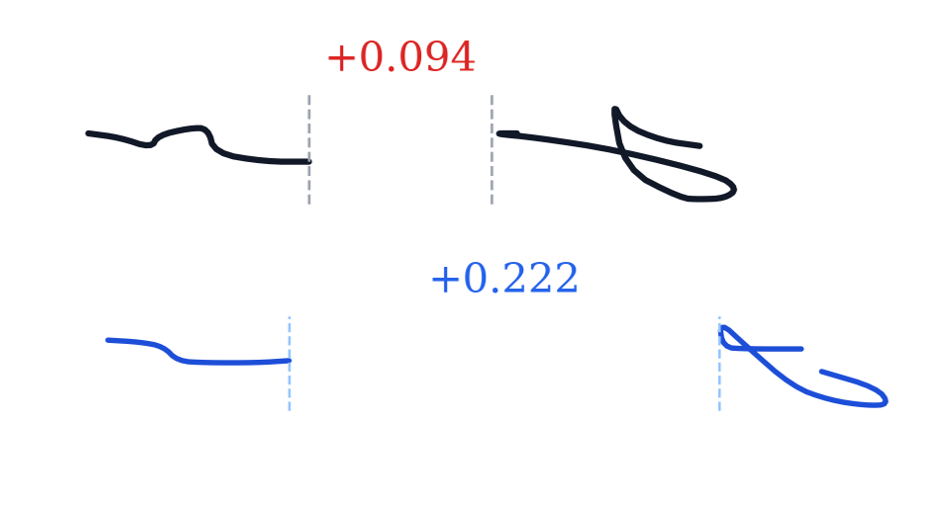}
    \caption{Illustrative example of SSS.
    SSS compares inter-word spacing widths using the same overlap-aware
    symmetric log-ratio similarity as KGS.}
    \label{fig:appendix_sss}
\end{figure}

Figure~\ref{fig:appendix_sss} illustrates the word-spacing comparison.
For the shown ground-truth and predicted widths of $0.094$ and $0.222$,
respectively, the resulting score is approximately $0.4234$.

KGS and SSS therefore evaluate different spatial relationships.
KGS operates on adjacent non-space character boundaries, whereas SSS
operates on gaps containing one or more whitespace characters.
Both are computed directly from trajectory geometry and do not depend on
whether a baseline predicts an explicit \CEOC\ state.

% ---------------------------------------------------------
% F.3 Aggregation and DTW Comparison
% ---------------------------------------------------------
\subsection{Writer-Macro Aggregation and DTW Comparison}
\label{sec:appendix_aggregation_dtw}

\paragraph{Writer-Macro Aggregation.}

The cursive metrics are defined directly at the writer level.
For the spacing metrics, we first average over all eligible boundaries for
each writer.

Writer-level KGS is
\begin{equation}
\mathrm{KGS}_w
=
\frac{
\displaystyle
\sum_{i\in\mathcal{D}_w}
\sum_{s\in\mathcal{B}_i}
\mathrm{KGS}^{(i)}_s
}{
\displaystyle
\sum_{i\in\mathcal{D}_w}
|\mathcal{B}_i|
+
\epsilon
},
\label{eq:kgs_writer}
\end{equation}
and writer-level SSS is
\begin{equation}
\mathrm{SSS}_w
=
\frac{
\displaystyle
\sum_{i\in\mathcal{D}_w}
\sum_{(u,v)\in\mathcal{P}_i}
\mathrm{SSS}^{(i)}_{u\rightarrow v}
}{
\displaystyle
\sum_{i\in\mathcal{D}_w}
|\mathcal{P}_i|
+
\epsilon
}.
\label{eq:sss_writer}
\end{equation}

For each component
$m\in
\{
\mathrm{F1}_{\mathrm{Cursive}},
\mathrm{CRE},
\mathrm{KGS},
\mathrm{SSS}
\}$,
the final writer-macro result is
\begin{equation}
m_{\mathrm{macro}}
=
\frac{1}{|\mathcal{W}|}
\sum_{w\in\mathcal{W}}
m_w.
\label{eq:appendix_csm_macro}
\end{equation}

A component is reported as undefined when the corresponding evaluation
protocol contains no eligible boundary type.
In particular, cursive-related metrics are undefined when a protocol has no
cursive-positive boundaries, and SSS is undefined when it contains no
whitespace runs.

\begin{figure}[t]
    \centering

    \begingroup
    \small
    \setlength{\tabcolsep}{3pt}

    % Three equal columns that exactly fill \columnwidth.
    \newlength{\csmCaseWidth}
    \setlength{\csmCaseWidth}{%
        \dimexpr
        (\columnwidth
         - 6\tabcolsep
         - 4\arrayrulewidth)/3
        \relax
    }

    % Identical vertical alignment for (a), (b), and (c).
    \newcommand{\csmHeaderCell}[1]{%
        \parbox[c][0.44cm][c]{\linewidth}{%
            \centering
            \bfseries #1%
        }%
    }

    % Full-width row title.
    \newcommand{\csmTitleRow}[1]{%
        \multicolumn{3}{|l|}{%
            \rule{0pt}{2.2ex}\textbf{#1}%
        }%
    }

    % Vertically and horizontally centered image cell.
    \newcommand{\csmImageCell}[2]{%
        \parbox[c][1.25cm][c]{\linewidth}{%
            \centering
            \includegraphics[
                width=0.94\linewidth,
                height=#2,
                keepaspectratio
            ]{#1}%
        }%
    }

    % Equal-width DeepWriting/Ours comparison.
    % The vertical separator is intentionally shorter than the cell height.
    \newcommand{\csmScoreCell}[2]{%
        \parbox[c][1.24cm][c]{\linewidth}{%
            \centering
            \makebox[\linewidth][c]{%
                \parbox[c][0.78cm][c]{0.43\linewidth}{%
                    \centering
                    \shortstack[c]{%
                        Deep\\
                        Writing\\
                        #1%
                    }%
                }%
                \hspace{0.025\linewidth}%
                \raisebox{-0.7ex}{%
                    \rule{0.35pt}{3.8ex}%
                }%
                \hspace{0.025\linewidth}%
                \parbox[c][0.78cm][c]{0.43\linewidth}{%
                    \centering
                    \shortstack[c]{%
                        Ours\\
                        #2%
                    }%
                }%
            }%
        }%
    }

    \begin{tabular}{
        |
        >{\centering\arraybackslash}m{\csmCaseWidth}
        |
        >{\centering\arraybackslash}m{\csmCaseWidth}
        |
        >{\centering\arraybackslash}m{\csmCaseWidth}
        |
    }
        \hline

        \csmHeaderCell{(a)}
        &
        \csmHeaderCell{(b)}
        &
        \csmHeaderCell{(c)}
        \\
        \hline

        % -------------------------------------------------
        % Ground Truth
        % -------------------------------------------------
        \csmTitleRow{Ground Truth}
        \\
        \hline

        \csmImageCell{
            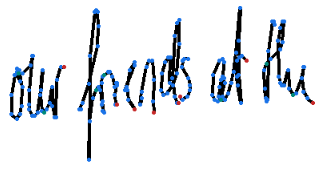
        }{1.18cm}
        &
        \csmImageCell{
            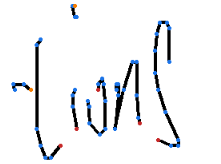
        }{1.18cm}
        &
        \csmImageCell{
            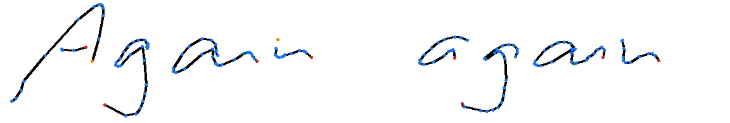
        }{1.18cm}
        \\
        \hline

        % -------------------------------------------------
        % DeepWriting
        % -------------------------------------------------
        \csmTitleRow{
            DeepWriting~\cite{aksan2018deepwriting}
        }
        \\
        \hline

        \csmImageCell{
            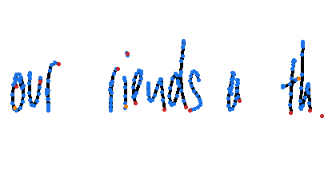
        }{1.18cm}
        &
        \csmImageCell{
            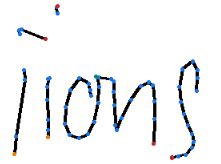
        }{1.18cm}
        &
        \csmImageCell{
            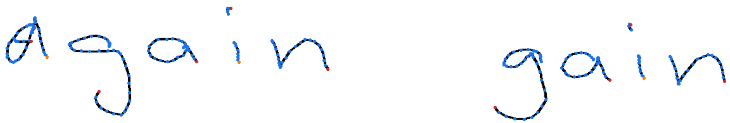
        }{1.18cm}
        \\
        \hline

        % -------------------------------------------------
        % Ours
        % -------------------------------------------------
        \csmTitleRow{Ours}
        \\
        \hline

        \csmImageCell{
            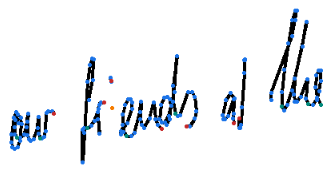
        }{1.18cm}
        &
        \csmImageCell{
            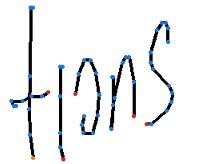
        }{1.18cm}
        &
        \csmImageCell{
            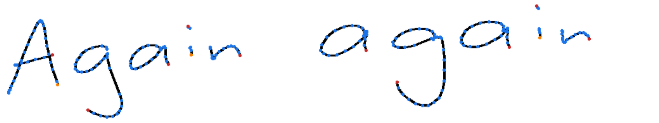
        }{1.18cm}
        \\
        \hline
        \hline

        % -------------------------------------------------
        % DTW: lower is better
        % -------------------------------------------------
        \csmTitleRow{DTW $\downarrow$}
        \\
        \hline

        \csmScoreCell{\textbf{0.153}}{0.171}
        &
        \csmScoreCell{\textbf{0.182}}{0.186}
        &
        \csmScoreCell{\textbf{0.437}}{0.513}
        \\
        \hline
        \hline

        % -------------------------------------------------
        % CSM mean: higher is better
        % -------------------------------------------------
        \csmTitleRow{CSM mean $\uparrow$}
        \\
        \hline

        \csmScoreCell{0.300}{\textbf{0.722}}
        &
        \csmScoreCell{0.339}{\textbf{0.714}}
        &
        \csmScoreCell{0.352}{\textbf{0.472}}
        \\
        \hline
    \end{tabular}

    \endgroup

    \caption{Representative cases of ranking disagreement between
    DTW~\cite{berndt1994using} and CSM.
    In (a), \textsc{DeepWriting}~\cite{aksan2018deepwriting} has a lower DTW
    than Ours ($0.153$ vs.\ $0.171$, $\Delta=0.018$), whereas Ours has a
    substantially higher CSM mean
    ($0.722$ vs.\ $0.300$, $\Delta=0.422$).
    In (b), DTW is nearly tied
    ($0.182$ vs.\ $0.186$, $\Delta=0.004$), but Ours achieves a considerably
    higher CSM mean
    ($0.714$ vs.\ $0.339$, $\Delta=0.375$).
    In (c), \textsc{DeepWriting} again has a lower DTW
    ($0.437$ vs.\ $0.513$, $\Delta=0.076$), while Ours retains a higher CSM
    mean
    ($0.472$ vs.\ $0.352$, $\Delta=0.120$).
    Lower DTW and higher CSM mean indicate better performance; bold denotes
    the better value for each metric.}
    \label{fig:appendix_csm_cases}
\end{figure}

\paragraph{Relationship to DTW.}

DTW measures the cost of globally aligning two trajectory sequences and is
therefore sensitive to overall path geometry, temporal progression, and
point-wise displacement.
However, a favorable DTW value does not necessarily indicate that individual
character boundaries exhibit the correct connection state or that local
character and word gaps follow the reference writer's spacing habits.
A generated trajectory may remain globally close to the reference while
locally altering a cursive join, introducing an unintended pen lift, or
compressing and expanding selected gaps.

CSM evaluates these boundary-local properties explicitly.
$\mathrm{F1}_{\mathrm{Cursive}}$ measures whether eligible character
boundaries are correctly classified as cursive-connected,
CRE compares the writer's overall tendency to use cursive continuation,
KGS evaluates adjacent-character gaps, and SSS evaluates inter-word spacing.
Consequently, CSM and DTW need not induce the same ranking because they
measure different aspects of generation quality.

Figure~\ref{fig:appendix_csm_cases} illustrates this distinction.
In cases~(a) and~(b), the DTW differences between the two methods are small,
yet the CSM differences are pronounced.
This indicates that similar global trajectory-alignment costs can coexist
with substantially different boundary connectivity and spacing behavior.
Case~(c) shows a stronger DTW preference for
\textsc{DeepWriting}~\cite{aksan2018deepwriting}, while CSM still favors
\textsc{BoundInk}.
The disagreement suggests that the globally closer trajectory does not
necessarily reproduce the reference writer's local boundary structure more
faithfully.

These examples should not be interpreted as evidence that CSM is generally
anti-correlated with DTW or that DTW is unsuitable for handwriting
evaluation.
They are selected qualitative cases that expose the different sensitivities
of the two measures.
DTW remains useful for evaluating global trajectory similarity, whereas CSM
provides a complementary assessment of the connectivity and spacing
properties targeted by \textsc{BoundInk}.
Reporting both therefore gives a more complete account of sentence-level
online handwriting quality than either metric alone.

\end{document}